\documentclass{article}

\usepackage{research_preprint}

\usepackage[utf8]{inputenc}
\usepackage[T1]{fontenc}
\usepackage{hyperref}
\usepackage{url}
\usepackage{booktabs}
\usepackage{graphicx}
\usepackage[ruled,vlined]{algorithm2e}
\usepackage{algorithmic}
\usepackage{amsmath}
\usepackage{amsfonts}
\usepackage{amssymb}
\usepackage{amsthm}
\usepackage{nicefrac}
\usepackage{microtype}
\usepackage{xcolor}

\title{Noise-Robust Conditional Flow Matching: Generating Clean Samples from Noisy Datasets}

\author{%
  Adrian Urbański\textsuperscript{1,2,3,*} \And
  Gabriel della Maggiora\textsuperscript{1,2,4,*} \And
  Artur Yakimovich\textsuperscript{1,2,3,5,\textsuperscript{+}} \\
  \\
  \textsuperscript{1}Center for Advanced Systems Understanding (CASUS), Görlitz, Germany \\
  \textsuperscript{2}Helmholtz-Zentrum Dresden-Rossendorf e. V. (HZDR), Dresden, Germany \\
  \textsuperscript{3}Institute of Computer Science, University of Wrocław, Wrocław, Poland \\
  \textsuperscript{4}School of Computation, Information and Technology, \\
  Technical University of Munich, Germany \\
  \textsuperscript{5}Cluster of Excellence Physics of Life, TU Dresden, Dresden, Germany \\
  \\
  \textbf{\textsuperscript{*}Equal contribution} \\
  \texttt{\textsuperscript{+}correspondance: \{a.yakimovich\}@hzdr.de}
}

\begin{document}

\maketitle

\begin{abstract}
Generative models learn the statistical properties of their training data, so
high-quality generation depends on clean and representative datasets. In
scientific imaging, acquisition often yields noisy measurements, while
collecting clean references can be costly, impractical or even unattainable.
Training directly on these measurements results in a model that reproduces the
corrupted data. This can be circumvented by learning the clean population distribution directly from the noisy data. Conditional flow matching (CFM) combines a simple regression objective with stable training, efficient sampling, and strong image-generation performance, making it a natural framework for this setting. We introduce Noise-Robust Conditional Flow Matching (NR-CFM), an unconditional generator
that learns from one corrupted observation per image. NR-CFM provides a
closed-form clean endpoint correction for additive white Gaussian noise and
learns a data-driven correction for general Gaussian corruptions with more complex covariance structure. Across the evaluated corruption
settings, NR-CFM outperforms NR-GAN in most cases and remains competitive with
Ambient Diffusion in the high-noise regime. We further evaluate NR-CFM on scientific data at signal-to-noise ratios as low as
$0.001$, where it generates plausible particle images from severely corrupted
measurements.
\end{abstract}

\twocolumn

\begin{figure*}[t]
  \centering
  \includegraphics[width=0.95\linewidth]{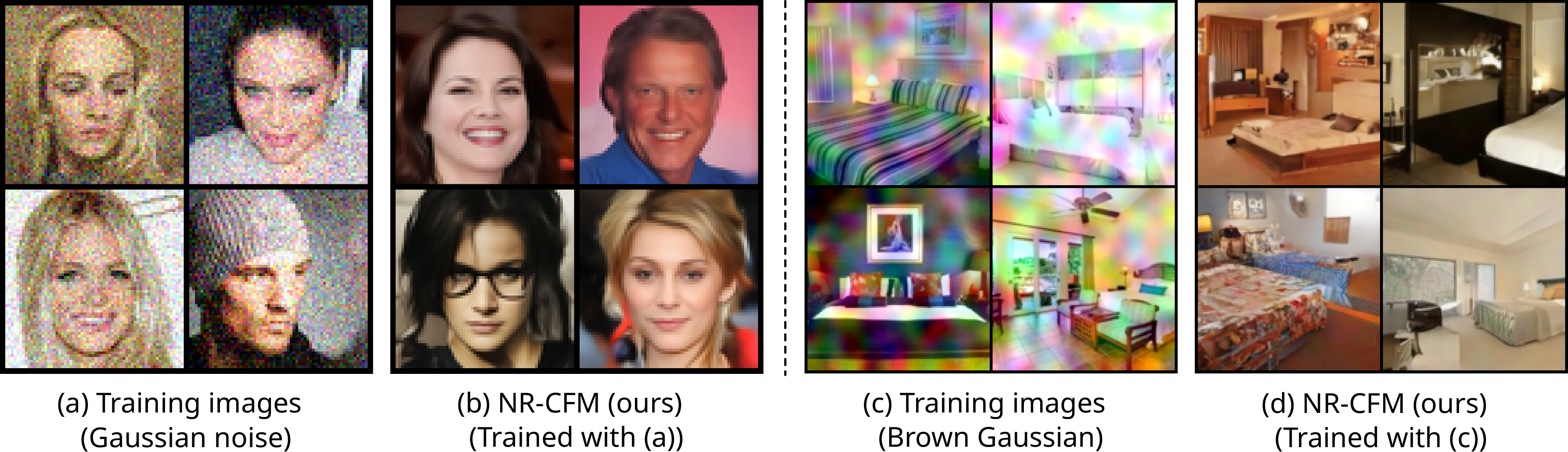}
  \caption{Unconditional generation from corrupted training observations with NR-CFM. The left panels show training images corrupted with additive Gaussian noise and Brown Gaussian noise. The right panels show samples generated from a Gaussian prior by models trained on the corresponding corrupted datasets. Training uses the corrupted observations together with the covariance or measurement metadata required by each corruption model.}
  \label{fig:main}
\end{figure*}

\section*{Introduction}
Scientific and computational imaging often seeks to characterize the
variability of structures, states, or phenotypes across a population.
Generative models provide a natural way to represent this variability by
learning a distribution from which new samples can be drawn. Building the
large, high-quality image collections required for this purpose is often
challenging, since acquisition pipelines may produce noisy or degraded
measurements, while improving image quality can require longer acquisition
times, greater radiation dose, or specialized instrumentation
\cite{kelkar2024ambientflow,kawar2024gsure}. Conditional flow matching (CFM)
has emerged as a powerful framework for generative modeling, achieving
state-of-the-art image generation through a stable regression-based training
objective and supporting flexible probability paths that enable fast and
reliable sampling \cite{lipman2023flow}. These properties make CFM a promising
basis for learning population distributions in scientific imaging. We develop
this framework for datasets composed of corrupted observations, with the goal
of learning an unconditional generator of the underlying clean-data
distribution.

Learning from corrupted data first became established in image restoration.
Noise2Noise showed that independently corrupted observations of the same image
can supervise a denoiser \cite{lehtinen2018noise2noise}. This principle was
subsequently extended to single noisy observations by Noise2Void and Noise2Self,
which construct self-supervised objectives from statistical structure across
pixels or measurement dimensions
\cite{krull2019noise2void,batson2019noise2self}. For Gaussian noise, SURE-based
objectives estimate the denoising risk directly from corrupted data
\cite{soltanayev2018sure}. These developments established that the supervision
required for image recovery can be extracted from the corrupted observations
themselves.

Generative modeling extends this principle from recovering individual images to
learning the full data distribution. AmbientGAN incorporated a known
measurement process into adversarial training
\cite{bora2018ambientgan}, while NR-GAN jointly modeled the image and noise
distributions under structural constraints \cite{kaneko2020noise}. Related
diffusion-based methods developed objectives for known measurement operators,
additive Gaussian corruption, and noisy linear measurements
\cite{daras2024ambientlaws,daras2024consistent,kawar2024gsure}. Flow-based
approaches have likewise learned invertible generators from incomplete
measurements and transports connecting observed and latent distributions
\cite{kelkar2024ambientflow,zhang2025inverse}. Together, these works show that
the structure of the acquisition process can provide sufficient information for
learning a clean generative distribution from corrupted datasets.

Building on this progression, we introduce Noise-Robust Conditional Flow
Matching (NR-CFM), an unconditional CFM framework trained from singly corrupted
observations. NR-CFM supports general Gaussian corruptions whose covariance may
vary across samples, contain spatial correlations, or depend on the signal.
This formulation encompasses fixed and variable AWGN, spatially structured
Gaussian noise, and signal-dependent Gaussian corruption. Training uses the
noise covariance for each sample, or enough measurement information to compute
and apply it, and sampling produces new clean-data samples from a Gaussian
prior.

Our detailed contributions are:
\begin{itemize}
    \item We derive an observed-bridge score identity and the associated
    posterior-mean corrections. The resulting correction is available in
    closed form for AWGN and extends to general Gaussian corruptions with
    sample-dependent, spatially structured, and signal-dependent covariance.

    \item Building on this identity, we introduce Noise-Robust Conditional Flow
    Matching (NR-CFM), which generates clean samples from a Gaussian prior after
    training only on corrupted observations.

    \item We evaluate NR-CFM across additive, spatially structured, and
    signal-dependent corruption regimes, including direct comparisons with
    NR-GAN and Ambient Diffusion and a low-SNR scientific-imaging stress test.
\end{itemize}
\section*{Background and Related Work}

Let $p_{\mathrm{data}}$ denote the latent clean-data distribution. We seek an
unconditional generative model for this distribution from singly corrupted
observations
\begin{equation}
    y=x_1+\varepsilon_y,
    \qquad
    \varepsilon_y\mid x_1,\omega
    \sim
    \mathcal{N}\!\left(0,\Sigma(x_1,\omega)\right),
    \label{eq:related_observation_model}
\end{equation}
where $x_1\sim p_{\mathrm{data}}$, $\omega$ identifies the noise setting for
that sample, and $\Sigma(x_1,\omega)\in\mathbb{R}^{d\times d}$ is positive
semidefinite. Each observation has Gaussian noise with covariance
$\Sigma(x_1,\omega)$. The covariance can change from one sample to another and
may depend on the clean image or on the sample's noise setting. We refer to this
family as general Gaussian corruption. Each latent clean datum contributes one
observed sample. During training, the noise covariance for the sample, or
measurement metadata sufficient to apply it, is available, while $x_1$ and the
actual noise draw remain unobserved.

\subsection*{Conditional Flow Matching}

Diffusion and score-based generative models represent a data distribution
through time-dependent denoising or score fields and generate samples by
reversing a prescribed corruption process
\citep{sohl2015deep,ho2020denoising,song2021score}. Flow Matching provides an
ODE-based route to the same broad goal. It trains a continuous normalizing flow
through simulation-free regression of the vector field associated with a
chosen family of conditional probability paths. Its Gaussian path family
contains standard diffusion paths, while optimal-transport paths provide
alternative couplings that produced efficient training and sampling in the
original experiments \citep{lipman2023flow}. Rectified flow and stochastic
interpolants developed closely related quadratic regression objectives for
learning continuous transports between distributions
\citep{liu2023flow,albergo2023building}.

In the standard CFM construction, one draws $x_0\sim\mathcal{N}(0,I)$ and
$x_1\sim p_{\mathrm{data}}$ and defines the linear conditional path
\begin{equation}
    x_t=(1-t)x_0+t x_1,
    \qquad t\in[0,1].
    \label{eq:related_linear_path}
\end{equation}
The path has samplewise velocity $x_1-x_0$, giving the regression objective of neural network $v_\theta$
\begin{equation}
    \mathcal{L}_{\mathrm{CFM}}(\theta)
    =
    \mathbb{E}_{t,x_0,x_1}
    \left[
        \left\|v_\theta(x_t,t)-(x_1-x_0)\right\|_2^2
    \right].
    \label{eq:cfm_loss}
\end{equation}
Its population minimizer $v^\star$ is the conditional mean velocity
\begin{equation}
    v_t^\star(x_t)
    =
    \mathbb{E}[x_1-x_0\mid x_t,t].
    \label{eq:related_cfm_velocity}
\end{equation}
Under the usual regularity conditions, this velocity defines the marginal
probability flow from the prior to $p_{\mathrm{data}}$
\citep{lipman2023flow,albergo2023building}. The construction uses samples from
the target endpoint distribution. In the setting of
\eqref{eq:related_observation_model}, the available endpoint is $y$, and the
clean endpoint $x_1$ remains latent.

\subsection*{Learning from Corrupted Observations}

Corrupted observations became a source of supervision through work on
self-supervised image restoration. Noise2Noise learns from independently
corrupted views of the same signal, while Noise2Void and Noise2Self obtain a
training signal from conditional independence across pixels or measurement
components. SURE-based objectives estimate denoising risk directly from noisy
observations under an explicit noise model
\citep{lehtinen2018noise2noise,krull2019noise2void,batson2019noise2self,soltanayev2018sure}.

Noisier2Noise introduced a re-corruption strategy for the setting in which each
latent image has only one noisy realization. Given a statistical model of the
noise, it applies an additional random corruption to the observed image and
trains a network to recover the original noisy observation from the noisier
input. The population regression can then be converted into a clean estimate.
The framework includes arbitrary additive noise, a Gaussian specialization,
multiplicative Bernoulli noise, and spatially structured corruption
\citep{moran2020noisier2noise}.
Recorrupted-to-Recorrupted develops a related construction in which synthetic
noisy views yield, under its assumptions, a loss equivalent in expectation to
supervised squared-error denoising \citep{pang2021recorrupted}.

A complementary line of work connects noisy-only learning to score estimation.
Noise2Score uses Tweedie's formula to express posterior image estimates through
the score of the noisy marginal and develops a common perspective on several
self-supervised denoising objectives. Its formulation covers Gaussian, Poisson,
and Gamma corruption within the exponential family \citep{kim2021noise2score}.
These results
are especially relevant to NR-CFM because the score of the observed bridge
links the learned flow to the clean posterior correction.

\subsection*{Generative Modeling from Corrupted Observations}

Generative modeling lifts corrupted-data learning from individual restoration
to the recovery of a population distribution. AmbientGAN trains a generator in
measurement space by applying a known stochastic measurement process to its
samples, with recovery governed by identifiability of that process
\citep{bora2018ambientgan}. NR-GAN jointly learns clean-image and noise
generators and uses distributional or transformation constraints to separate
the two components under incomplete noise information. Its variants cover both
signal-independent and signal-dependent corruption \citep{kaneko2020noise}.

Score- and diffusion-based methods use several forms of corrupted-only
supervision. SURE-Score learns an approximate clean score prior from additive
Gaussian observations \citep{aali2023surescore}. GSURE-based diffusion training extends
Stein risk estimation to noisy and incomplete linear measurements under
conditions that connect the objective to supervised diffusion training
\citep{kawar2024gsure}. Ambient Diffusion uses deliberate further corruption at
the distribution level: it adds measurement distortion to a corrupted training
example and predicts the original corrupted image from the further-corrupted
input. Under its identifiability conditions, this learns a clean conditional
expectation for measurement operators including inpainting and compressed
sensing \citep{daras2024ambientlaws}. Daras et al. subsequently combined a
double application of Tweedie's formula with a consistency objective to obtain
provably exact sampling from the clean distribution in the additive Gaussian
setting \citep{daras2024consistent}.

The deliberate re-corruption used by Noisier2Noise and Ambient Diffusion is
also the high-level principle behind the NR-CFM auxiliary correction. The three
methods apply it to different learned quantities: Noisier2Noise estimates an
individual clean image, Ambient Diffusion learns a clean generative
distribution, and NR-CFM identifies a correction field on the observed CFM
bridge. In NR-CFM, the added perturbation is tied to the realized conditional
covariance so that the auxiliary regression retains the sample-specific and
state-dependent noise information.

Iterative approaches provide another route to clean-distribution learning.
Expectation-maximization methods alternate posterior reconstruction of latent
clean samples with updates of a diffusion prior from noisy or incomplete
observations \citep{bai2024expectation}. A related expectation-maximization
formulation learns diffusion priors directly from observations
\citep{rozet2024learning}.
A subsequent scaling study quantifies the finite-sample cost of training
ambient diffusion models from noisy images \citep{daras2024much}. SFBD studies
the same finite-data difficulty through deconvolution theory and uses limited
clean data or related pretraining to guide the learned diffusion model
\citep{lu2025stochastic}.

Transport-based models have developed parallel formulations. AmbientFlow uses
variational Bayesian learning to fit an invertible generative model directly
from noisy and incomplete measurements \citep{kelkar2024ambientflow}. Inverse
Flow learns a transport from an observed noisy distribution to a latent clean
distribution and supports broad continuous noise models with complex
dependencies \citep{zhang2025inverse}. Self-consistent stochastic interpolants
iteratively learn a clean transport from corrupted observations using black-box
access to a potentially nonlinear corruption channel \citep{modi2026scsi}.

NR-CFM develops an unconditional CFM generator for the general Gaussian
corruption model in \eqref{eq:related_observation_model}. Sampling begins from
a Gaussian prior. The noise covariance can vary across samples and represent
spatially structured or signal-dependent noise. AWGN is a special case of this
model.

\section*{Methods}
\label{sec:methods}

We consider training data in which the clean sample is never observed directly.
Let $x_1\sim p_{\mathrm{data}}$ denote the clean endpoint.  Each sample is
observed once through a Gaussian corruption,
\begin{equation}
    y=x_1+\varepsilon_y,
    \qquad
    \varepsilon_y\mid x_1,\omega
    \sim
    \mathcal{N}\!\left(0,\Sigma(x_1,\omega)\right).
    \label{eq:observation_model}
\end{equation}
We use the general Gaussian corruption model in
\eqref{eq:observation_model}. Here, $\omega$ identifies the sample's noise
setting. The covariance can vary across samples and represent spatially
structured or signal-dependent noise.

As in standard CFM, we draw $x_0\sim\mathcal{N}(0,I)$ independently of the data
and the corruption.  If the clean endpoint were available, the corresponding
bridge would be
\[
    x_t=(1-t)x_0+t x_1.
\]
Since only $y$ is observed, we instead train on the observed bridge
\begin{equation}
    y_t=(1-t)x_0+ty=x_t+t\varepsilon_y .
    \label{eq:bridges}
\end{equation}
This bridge starts from the same Gaussian prior as the clean bridge but ends at
the corrupted data distribution.  NR-CFM first learns its velocity field, then
uses that field to estimate the observed-bridge score and correct the state and
velocity toward the clean bridge.  Section~\ref{sec:observed_bridge_flow}
constructs the score estimate, Sections~\ref{sec:awgn_correction}
and~\ref{sec:hidden_mode_correction} give the corrections for AWGN and general
Gaussian corruption, and Section~\ref{sec:sample_generation} describes
sample generation.

\subsection*{Flow Matching on the Observed Bridge}
\label{sec:observed_bridge_flow}

Consider a CFM model trained directly on the noisy observations.
Because $y$ is available during training, the observed bridge in
\eqref{eq:bridges} can be treated as an ordinary conditional flow-matching path.
For each sampled pair $(x_0,y)$, the path has the constant samplewise velocity
$y-x_0$.  We therefore train
\begin{equation}
    \begin{aligned}
        \mathcal{L}_{\mathrm{obs}}(\theta)
        &=
        \mathbb{E}_{t,x_0,y}
        \left[
            \left\|
            v_\theta(y_t,t)-(y-x_0)
            \right\|^2
        \right], \\
        v_t^\star(y_t)
        &=
        \mathbb{E}[y-x_0\mid y_t,t].
    \end{aligned}
    \label{eq:observed_cfm_loss}
\end{equation}
Under the squared loss, $v_t^\star$ is the population minimizer and the marginal
velocity field of the observed bridge, whose endpoint distribution is the law
of $y$ (see Appx.~\ref{app:proof_observed_flow}).  Consequently, we use
$v_t^\star$ in the population-level analysis below and replace it with its
learned approximation $v_\theta$ in practice.

The population velocity also determines the exact score of the observed
bridge.  Let $p_t^{\mathrm{noisy}}$ denote the density of $y_t$.  The Gaussian
prior $x_0$ enters the bridge with scale $1-t$.  Applying the Gaussian Tweedie
identity to this term and using the definition of $v_t^\star$ in
\eqref{eq:observed_cfm_loss} gives
\begin{equation}
    \nabla_{y_t}\log p_t^{\mathrm{noisy}}(y_t)
    =
    \frac{t\,v_t^\star(y_t)-y_t}{1-t},
    \qquad t\in(0,1).
    \label{eq:score_from_velocity}
\end{equation}
The full derivation is given in Appx~\ref{app:proof_score_awgn}.
Because $v_t^\star$ is not available directly, we replace it in practice by the
learned model $v_\theta$ and define the score estimate
\begin{equation}
    g_\theta(y_t,t)
    :=
    \frac{t\,v_\theta(y_t,t)-y_t}{1-t}.
    \label{eq:score_anchor}
\end{equation}
We use $g_\theta$ as the score estimate in the correction terms below.  Since
the denominator vanishes at $t=1$, the estimate is evaluated only before the
terminal endpoint; Section~\ref{sec:sample_generation} describes the
corresponding cutoff and endpoint readout.

\subsection*{Closed-Form Correction for Additive White Gaussian Noise}
\label{sec:awgn_correction}

We first consider additive white Gaussian noise (AWGN),
\[
    \varepsilon_y\sim\mathcal{N}(0,\sigma^2I).
\]
Recall from \eqref{eq:bridges} that $y_t=x_t+t\varepsilon_y$.  Therefore, at
time $t$, the observed bridge is the clean bridge blurred by Gaussian noise:
\begin{equation}
    p_t^{\mathrm{noisy}}
    =
    p_t^{\mathrm{clean}}
    *
    \mathcal{N}\!\left(0,(\sigma t)^2I\right).
    \label{eq:awgn_noisy_marginal}
\end{equation}
The blur is absent at $t=0$ and reaches the full observation noise at $t=1$.
The corresponding marginal dynamics are derived in
Appx.~\ref{app:proof_path_dynamics}.

To recover clean quantities from $y_t$, we use the conditional mean of the
observation noise.  Since the noise enters the bridge multiplied by $t$, define
\begin{equation}
    s_t(y_t)
    :=
    -\frac{1}{t}\,
    \mathbb{E}[\varepsilon_y\mid y_t,t],
    \qquad t>0.
    \label{eq:correction_field}
\end{equation}
This gives the posterior mean clean state and clean velocity:
\begin{equation}
    \begin{aligned}
        \mathbb{E}[x_t\mid y_t,t]
        &=
        y_t+t^2s_t(y_t), \\
        \mathbb{E}[x_1-x_0\mid y_t,t]
        &=
        v_t^\star(y_t)+t\,s_t(y_t).
    \end{aligned}
    \label{eq:clean_state_velocity_from_correction}
\end{equation}
Thus, $t^2s_t$ corrects the bridge state, while $t\,s_t$ corrects the observed
velocity.

For AWGN, Tweedie's formula gives
\begin{equation}
    \begin{aligned}
        s_t(y_t)
        &=
        \sigma^2\nabla_{y_t}\log p_t^{\mathrm{noisy}}(y_t) \\
        &=
        \sigma^2\frac{t\,v_t^\star(y_t)-y_t}{1-t},
        \qquad t\in(0,1).
    \end{aligned}
    \label{eq:awgn_correction_field}
\end{equation}
Using the score estimate $g_\theta$ from \eqref{eq:score_anchor}, the practical
correction is
\begin{equation}
    \widehat{s}(y_t,t)
    =
    \sigma^2g_\theta(y_t,t).
    \label{eq:awgn_practical_correction}
\end{equation}
Thus, for AWGN, the learned velocity $v_\theta$ directly determines the
practical correction and hence the clean-state and clean-velocity estimates.
At $t=0$, the
products in \eqref{eq:clean_state_velocity_from_correction} are interpreted by
continuity and both correction terms vanish.  The full derivation is given in
Appx.~\ref{app:proof_score_awgn}.

\subsection*{General Gaussian Correction}
\label{sec:hidden_mode_correction}

We now consider general Gaussian corruption. Let
\[
    S:=\Sigma(x_1,\omega)
\]
be the noise covariance for the current sample. The correction depends on the
average noise covariance at the current bridge state,
\begin{equation}
    \bar{\Sigma}_t(y_t)
    :=
    \mathbb{E}[S\mid y_t,t],
    \label{eq:posterior_expected_covariance}
\end{equation}

Under the regularity assumptions in
Appx.~\ref{app:proof_hidden_mode_gaussian}, the correction field is
\begin{equation}
    s_t(y_t)
    =
    \bar{\Sigma}_t(y_t)
    \nabla_{y_t}\log p_t^{\mathrm{noisy}}(y_t)
    +
    \operatorname{div}\bar{\Sigma}_t(y_t).
    \label{eq:general_correction_field}
\end{equation}
The divergence term appears because the average noise covariance can change
with the bridge state.
Using \eqref{eq:score_from_velocity}, the same expression can be written as
\[
    \begin{aligned}
        s_t(y_t)
        &=
        \bar{\Sigma}_t(y_t)g_t^\star(y_t)
        +
        \operatorname{div}\bar{\Sigma}_t(y_t), \\
        g_t^\star(y_t)
        &:=
        \frac{t\,v_t^\star(y_t)-y_t}{1-t}.
    \end{aligned}
\]
For AWGN, every sample has covariance $S=\sigma^2I$. Therefore,
$\bar{\Sigma}_t(y_t)=\sigma^2I$ and the divergence term vanishes. For general
Gaussian corruption, the covariance varies across samples and bridge states,
so we estimate the correction with an auxiliary network $s_\phi$.

Each observation must include its covariance
$S_i=\Sigma(x_{1,i},\omega_i)$, or sufficient measurement metadata to apply
$S_i^{1/2}$; neither the clean image nor the realized noise is required.

For a sample with covariance $S$, draw $\eta\sim\mathcal{N}(0,I)$ and further
perturb the observed bridge as
\begin{equation}
    \Delta_{t,\delta}
    =
    \sqrt{\delta}\,t\,S^{1/2}\eta,
    \qquad
    \widetilde{y}_{t,\kappa,\delta}
    =
    y_t+\kappa\Delta_{t,\delta},
    \label{eq:further_corruption}
\end{equation}
where $\delta>0$, $\kappa\neq0$, and $t>0$. Reusing the same $S$ preserves the
sample-specific covariance information needed for the correction.

We parameterize the general Gaussian correction as
\begin{equation}
    s_\phi(y_t,t)
    =
    A_\phi(y_t,t)\operatorname{sg}\!\left[g_\theta(y_t,t)\right]
    +
    r_\phi(y_t,t),
    \label{eq:correction_param}
\end{equation}
where $\operatorname{sg}$ denotes stop-gradient. We implement $A_\phi$ as an
elementwise multiplier, giving a practical diagonal covariance approximation,
while $r_\phi$ captures the remaining covariance effects and the divergence
term in \eqref{eq:general_correction_field}.

The auxiliary loss regresses against the known added perturbation:
\begin{equation}
    \mathcal{L}_{\mathrm{aux}}(\phi)
    =
    \mathbb{E}
    \left[
        \left\|
        -\kappa\delta t^2
        s_\phi(\widetilde{y}_{t,\kappa,\delta},t)
        -
        \Delta_{t,\delta}
        \right\|^2
    \right].
    \label{eq:auxiliary_loss}
\end{equation}
For finite $\delta$, this loss identifies the correction of the additionally
smoothed bridge; under the assumptions in
Appx.~\ref{app:proof_auxiliary_correction}, it converges to the desired field
as $\delta\rightarrow0$. Appx.~\ref{subsec:ablations} studies sensitivity to
$\delta$ and $\kappa$. The combined objective is
\begin{equation}
        \mathcal{L}(\theta,\phi)
        =
        \mathcal{L}_{\mathrm{obs}}(\theta)
        +
        \lambda_{\mathrm{aux}}\mathcal{L}_{\mathrm{aux}}(\phi).
    \label{eq:total_loss}
\end{equation}

\subsection*{Sample Generation}
\label{sec:sample_generation}

To generate clean samples, NR-CFM transports a Gaussian prior sample along the
learned observed bridge and then maps the resulting bridge state to a clean
endpoint estimate.  For a bridge state $y_t$, the endpoint map uses the learned
correction
\begin{equation}
    \widehat{s}(y_t,t)
    =
    \begin{cases}
        \sigma^2 g_\theta(y_t,t), & \text{AWGN},\\
        s_\phi(y_t,t), & \text{general Gaussian}.
    \end{cases}
    \label{eq:sampler_correction_choice}
\end{equation}
Combining the learned clean-state and clean-velocity estimates gives the
endpoint map
\begin{equation}
    \begin{aligned}
        \widehat{m}_t(y_t)
        &:={}
        y_t+t^2\widehat{s}(y_t,t) \\
        &\quad+
        (1-t)\bigl(v_\theta(y_t,t)+t\widehat{s}(y_t,t)\bigr).
    \end{aligned}
    \label{eq:endpoint_proposal}
\end{equation}

\begin{algorithm}[t]
\caption{NR-CFM sample generation}
\label{alg:nr_cfm_sampling}
\begin{algorithmic}[1]
\REQUIRE Trained velocity $v_\theta$, correction $\widehat{s}$, and an ODE
solver with $N$ steps
\STATE Set $t_{\mathrm{cut}}=0.95$, for which
$(1-t_{\mathrm{cut}})^{-1}=20$
\STATE Draw $x_0\sim\mathcal{N}(0,I)$
\STATE Integrate $d y_t/dt=v_\theta(y_t,t)$ from $y_0=x_0$ to
$t=t_{\mathrm{cut}}$, obtaining $y_{t_{\mathrm{cut}}}$
\RETURN $\widehat{x}_1=
\widehat{m}_{t_{\mathrm{cut}}}(y_{t_{\mathrm{cut}}})$
\end{algorithmic}
\end{algorithm}

Only the terminal endpoint readout is returned; intermediate readouts are not
used.

The correction field in \eqref{eq:sampler_correction_choice} contains the score
factor $(1-t)^{-1}$ inherited from \eqref{eq:score_from_velocity}, which grows
rapidly as $t\to1$.  Pushing the corrected quantities all the way to $1$ would
therefore amplify estimation errors and destabilize the endpoint prediction.
The cutoff in Algorithm~\ref{alg:nr_cfm_sampling} therefore evaluates
\eqref{eq:endpoint_proposal} before the terminal singularity and uses the
endpoint map for the final jump to time $1$.
Appx.~\ref{subsec:ablations} evaluates the sensitivity of generation quality
to the choice of $t_{\mathrm{cut}}$.

\begin{figure}[!tb]
\centering
\includegraphics[width=\linewidth]{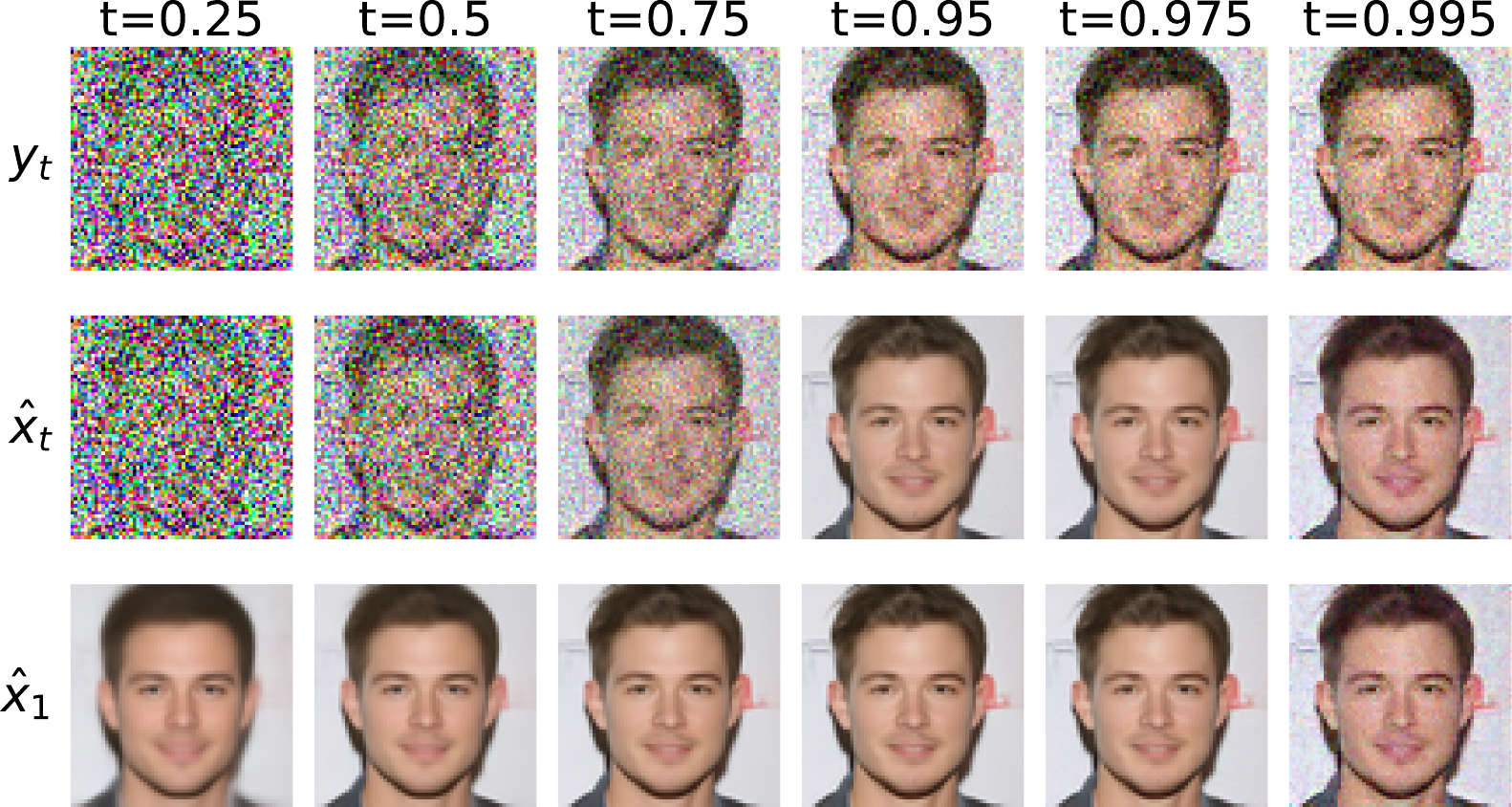}\caption{Sample trajectories for NR-CFM on CelebA-HQ under AWGN corruption. Here $y_t$ denotes the noisy bridge state, $\widehat{x}_t$ its cleaned estimate at time $t$, and $\widehat{x}_1$ the final clean endpoint prediction. Corruption is most visible near $t=1$ because the score term is scaled by $(1-t)^{-1}$, which motivates stopping generation at an earlier point.}

\label{fig:celeba-path}
\end{figure}

To characterize the distribution sampled by
Algorithm~\ref{alg:nr_cfm_sampling}, recall that $p_t^{\mathrm{noisy}}$ denotes
the observed-bridge distribution, with random state
$Y_t=(1-t)X_0+tY$; lowercase $y$ denotes a realized bridge state.  For
$t\in(0,1)$, define the population clean endpoint readout
\begin{equation}
    \begin{aligned}
    m_t(y)
    &:={}
    y+t^2s_t(y)
    +(1-t)\bigl(v_t^\star(y)+t\,s_t(y)\bigr) \\
    &=
    \mathbb{E}[X_1\mid Y_t=y,t].
    \end{aligned}
    \label{eq:population_endpoint_readout}
\end{equation}
The first two terms are the posterior mean clean bridge state, while the term
in parentheses is the posterior mean clean bridge velocity.  The second line
follows from $X_t+(1-t)(X_1-X_0)=X_1$.  At the population optimum,
$\widehat{m}_t(y)=m_t(y)$.

Passing the observed bridge distribution through this endpoint map gives the
generated clean distribution
\begin{equation}
    q_t
    :=
    (m_t)_{\#}p_t^{\mathrm{noisy}}.
    \label{eq:generated_clean_distribution}
\end{equation}
At the population optimum, the observed-bridge dynamics produce
$Y_t\sim p_t^{\mathrm{noisy}}$.  Algorithm~\ref{alg:nr_cfm_sampling} is
therefore the learned approximation to sampling from the pushforward
$q_{t_{\mathrm{cut}}}$.  The target generative relation is
$q_{t_{\mathrm{cut}}}\approx p_{\mathrm{data}}$, which we evaluate empirically
using FID against clean reference statistics.

At the terminal endpoint, the posterior-mean identity extends the readout
to $t=1$ as
\begin{equation}
    \begin{aligned}
        p_1^{\mathrm{noisy}}
        &= p_Y, \\
        m_1(y)
        &= \mathbb{E}[X_1\mid Y=y], \\
        q_1
        &= (m_1)_{\#}p_Y,
    \end{aligned}
    \label{eq:terminal_generated_distribution}
\end{equation}
where $p_Y$ is the law of the corrupted endpoint.  The reported implementation
targets the cutoff distribution $q_{t_{\mathrm{cut}}}$ for numerical stability,
instead of evaluating the readout at $t=1$.
Fig.~\ref{fig:celeba-path} illustrates the resulting NR-CFM sampling
trajectory on CelebA-HQ, from the noisy bridge state to the final clean endpoint
readout.

\section*{Experimental Setup}
\label{sec:experimental_setup}
We compare noise-robust conditional flow matching (NR-CFM) with NR-GAN across
Gaussian corruptions \cite{kaneko2020noise} and Ambient Diffusion under
additive white Gaussian noise (AWGN)
\cite{daras2024ambientlaws,daras2024much}, following their evaluation protocols
and published results. CryoBench provides a low-SNR scientific-imaging stress
test \cite{jeon2024cryobench}.

Our primary metric is Fr\'echet Inception Distance (FID; lower is better).
Because the source protocols differ, values are comparable only within each
result table. Dataset, corruption, evaluation, training, and sampling details, as well as exact evaluation protocols for each experiment,
are provided in Appx.~\ref{app:implementation_details}.

\paragraph{NR-GAN comparison.}
We evaluate CIFAR-10 \cite{krizhevsky2009learning}, LSUN Bedroom
\cite{yu2015lsun}, and FFHQ \cite{karras2019style} on the Gaussian subset of
the NR-GAN benchmark \cite{kaneko2020noise}. We use settings (A)--(D) and
(G)--(L) on CIFAR-10, A, B, G, I, and L on LSUN Bedroom, and A and I on FFHQ.
These corruption settings are Gaussian conditional on the clean image and a hidden corruption state,
matching our assumptions.
Published NR-GAN and AmbientGAN results serve as baselines; AmbientGAN receives
the ground-truth corruption model. Full benchmark and corruption specifications
are provided in Appx.~\ref{app:nrgan_details}.

\paragraph{Ambient Diffusion comparison.}
We compare with Ambient Diffusion as a method specialized to known AWGN. We use
CIFAR-10 and CelebA-HQ \cite{liu2015deep} in the setup of
\cite{daras2024ambientlaws,daras2024much} across several AWGN levels. We report
the published values from the scaling study of Daras et al.~\cite{daras2024much}; comprehensive evaluation details are provided in
Appx.~\ref{app:ambient_diffusion_details}.

\paragraph{CryoBench evaluation.}
For the scientific-imaging stress test, we use the synthetic IgG-1D dataset from
CryoBench \cite{jeon2024cryobench}, a benchmark suite for heterogeneous cryo-EM
reconstruction. It provides simulated image stacks at multiple signal-to-noise
ratios along a controlled one-dimensional conformational trajectory. We use it
only as an extreme-noise generation test, not to evaluate structural inference
or cryo-EM reconstruction. Dataset and image-formation details are given in
Appx.~\ref{app:cryobench_details}.
\section*{Results}
\label{sec:results}
We evaluate NR-CFM against NR-GAN on CIFAR-10, LSUN Bedroom, and FFHQ, against Ambient Diffusion under known AWGN, and on the
extreme-noise CryoBench setting. NR-CFM improves over NR-GAN on most Gaussian
corruptions and becomes competitive with Ambient Diffusion as noise increases.
For context, we also report vanilla CFM performance to isolate the gains from
our noise-robust correction. Tabs.~\ref{tab:nrgan-cifar}
and~\ref{tab:nrgan-lsun-ffhq} additionally include clean-data CFM performance in their captions
as an empirical lower bound on FID.

\paragraph{NR-GAN benchmark comparisons.}
Tabs.~\ref{tab:nrgan-cifar} and~\ref{tab:nrgan-lsun-ffhq} compare NR-CFM with NR-GAN on the unknown-corruption benchmarks reported by \cite{kaneko2020noise}. Tab.~\ref{tab:nrgan-cifar} reproduces the CIFAR-10 FID benchmark for the corruption settings included in our main comparison. These settings span both signal-independent and signal-dependent corruptions and therefore provide a broad stress test for unconditional generation from fully corrupted observations. NR-CFM improves over the best reported NR-GAN variant by a clear margin on every setting shown.

Tab.~\ref{tab:nrgan-lsun-ffhq} extends the comparison to the larger-image LSUN Bedroom and FFHQ benchmarks. On LSUN Bedroom, NR-CFM gives the best FID for the signal-independent settings A, B, and G. Its performance is weaker on the multiplicative and additive-plus-multiplicative settings I and L, where the published NR-GAN and AmbientGAN numbers remain lower. On FFHQ, NR-CFM improves over or matches NR-GAN and AmbientGAN on both reported settings. These results show that the gains on CIFAR-10 transfer to several larger-image settings, but also identify signal-dependent LSUN corruptions as a current limitation.

\begin{table*}[t]
\centering
\caption{FID ($\downarrow$) on CIFAR-10 under the NR-GAN corruption settings. Setting labels and corruption abbreviations follow the notation of \cite{kaneko2020noise}. ``SI/SD-NR-GAN'' denotes the best model in each family; $*$ indicates that NR-GAN does not use the ground-truth corruption model. The FID of a CFM model trained on clean data is 8.35.}
\label{tab:nrgan-cifar}
\resizebox{\linewidth}{!}{%
\begin{tabular}{lcccccccccc}
\toprule
& \multicolumn{6}{c}{\textit{Signal-independent}}
& \multicolumn{4}{c}{\textit{Signal-dependent}} \\
\cmidrule(lr){2-7}\cmidrule(lr){8-11}
\textit{Setting} & A & B & C & D & G & H & I & J & K & L \\
\textit{Corruption} & AGF & AGV & LGF & LGV & BG & A+G
& MGF & MGV & A+I & A+I \\
\midrule
CFM & 135.0 & 119.2 & 26.4 & 26.4 & 163.6 & 209.2
& 54.7 & 48.8 & 51.1 & 137.2 \\
AmbientGAN & 26.7 & 28.0 & 21.8 & 21.7 & 30.3 & 40.8
& 21.4 & 21.8 & 21.9 & 27.4 \\
SI/SD-NR-GAN$*$ & 26.7 & 27.5 & 22.1 & 21.7 & 32.2 & 44.0
& 22.5 & 23.0 & 23.3 & 28.5 \\
NR-CFM (ours) & \textbf{17.4} & \textbf{12.8} & \textbf{10.3}
& \textbf{10.0} & \textbf{15.0} & \textbf{18.0} & \textbf{11.6} & \textbf{9.8} & \textbf{10.1} & \textbf{13.8} \\
\bottomrule
\end{tabular}}
\end{table*}

\begin{table}[t]
\centering
\caption{FID ($\downarrow$) on LSUN Bedroom and FFHQ. Setting labels and corruption abbreviations follow the notation of \cite{kaneko2020noise}.
Bold denotes the best result and underline denotes results within 1 FID.
``SI/SD-NR-GAN'' reports the best model in each family; $*$ indicates no ground-truth corruption model. The FID of a CFM model trained on clean data is 8.6 for LSUN Bedroom and 16.9 for FFHQ.}
\label{tab:nrgan-lsun-ffhq}
\scriptsize
\resizebox{\linewidth}{!}{%
\begin{tabular}{lccccccc}
\toprule
& \multicolumn{4}{c}{Signal-independent} & \multicolumn{3}{c}{Signal-dependent} \\
\cmidrule(lr){2-5}\cmidrule(lr){6-8}
& \multicolumn{3}{c}{LSUN Bedroom} & FFHQ & \multicolumn{2}{c}{LSUN Bedroom} & FFHQ \\
\cmidrule(lr){2-4}\cmidrule(lr){5-5}\cmidrule(lr){6-7}\cmidrule(lr){8-8}
Method & A & B & G & A & I & L & I \\
\textit{Corruptions} & AGF & AGV & BG & AGF & MGF & A+I & MGF \\
\midrule
CFM & 107.3 & 108.8 & 148.8 & 85.5 & 79.4 & 114.4 & 61.8\\
AmbientGAN & 19.4 & 25.0 & \underline{9.7} & \underline{28.3} & \underline{11.7} & 19.2 & \textbf{18.7} \\
SI/SD-NR-GAN$*$ & 13.8 & \underline{14.2} & 10.8 & 35.7 & \textbf{11.6} & \textbf{15.0} & 26.5 \\
NR-CFM (ours) & \textbf{12.4} & \textbf{13.3} & \textbf{9.2} & \textbf{27.9} & 28.4 & 51.1 & \textbf{18.7} \\
\bottomrule
\end{tabular}}
\end{table}

\paragraph{AWGN comparison against ambient diffusion.}
The known-corruption setting is the regime where ambient-diffusion methods are strongest.
To position NR-CFM relative to that line of work, Tab.~\ref{tab:ambient-awgn} summarizes the AWGN results reported in the recent ambient-diffusion scaling study of Daras et al.
\cite{daras2024much}.
We include noise standard deviations $\sigma\in\{0,0.05,0.1,0.2\}$ for CIFAR-10 and CelebA-HQ.
Following the source study, we use the full-sampling ambient-diffusion result for $\sigma=0$ and the consistency-plus-full-sampling result for nonzero noise levels.
Ambient Diffusion outperforms NR-CFM on most entries, especially in the
low-corruption regime: at $\sigma=0$, the FID gap is roughly two points on
CIFAR-10 and two and a half points on CelebA-HQ.  The gap narrows as corruption
increases, and at $\sigma=0.2$, NR-CFM improves over the reported Ambient
Diffusion result on both datasets.

\begin{table}[t]
\centering
\caption{
FID ($\downarrow$) on CIFAR-10 and CelebA-HQ under AWGN.
Bold denotes the best result and underline denotes results within 1 FID.
Values are mean $\pm$ standard deviation.
}
\label{tab:ambient-awgn}
\scriptsize
\setlength{\tabcolsep}{2pt}
\resizebox{\linewidth}{!}{%
\begin{tabular}{@{}lcccc@{}}
\toprule
\multicolumn{5}{c}{CIFAR-10} \\
\cmidrule(lr){1-5}
Method / noise std. $\sigma$ & 0.00 & 0.05 & 0.10 & 0.20 \\
\midrule
CFM & 3.97 $\pm$ 0.01 & 44.55 $\pm$ 0.12 & 82.98 $\pm$ 0.08
& 135.98 $\pm$ 0.18 \\
Ambient diffusion & \textbf{1.99 $\pm$ 0.02} & \textbf{2.82 $\pm$ 0.02} & \textbf{3.63 $\pm$ 0.03} & 11.93 $\pm$ 0.09 \\
NR-CFM (ours) & 3.97 $\pm$ 0.01 & 7.00 $\pm$ 0.04
& 9.51 $\pm$ 0.04 & \textbf{9.71 $\pm$ 0.12} \\
\bottomrule
\end{tabular}}

\medskip

\resizebox{\linewidth}{!}{%
\begin{tabular}{@{}lcccc@{}}
\toprule
\multicolumn{5}{c}{CelebA-HQ} \\
\cmidrule(lr){1-5}
Method / noise std. $\sigma$ & 0.00 & 0.05 & 0.10 & 0.20 \\
\midrule
CFM & 4.93 $\pm$ 0.06 & 40.09 $\pm$ 0.15 & 62.81 $\pm$ 0.02 & 108.32 $\pm$ 0.08 \\
Ambient diffusion & \textbf{2.40 $\pm$ 0.07} & \textbf{5.50 $\pm$ 0.03} & \textbf{9.38 $\pm$ 0.02} & 12.97 $\pm$ 0.11 \\
NR-CFM (ours) & 4.93 $\pm$ 0.06 & 7.15 $\pm$ 0.03 & \underline{9.88 $\pm$ 0.05} & \textbf{10.09 $\pm$ 0.09} \\
\bottomrule
\end{tabular}}
\end{table}

\paragraph{CryoBench dataset}

\begin{figure}[t]
\centering
\includegraphics[width=0.95\linewidth]{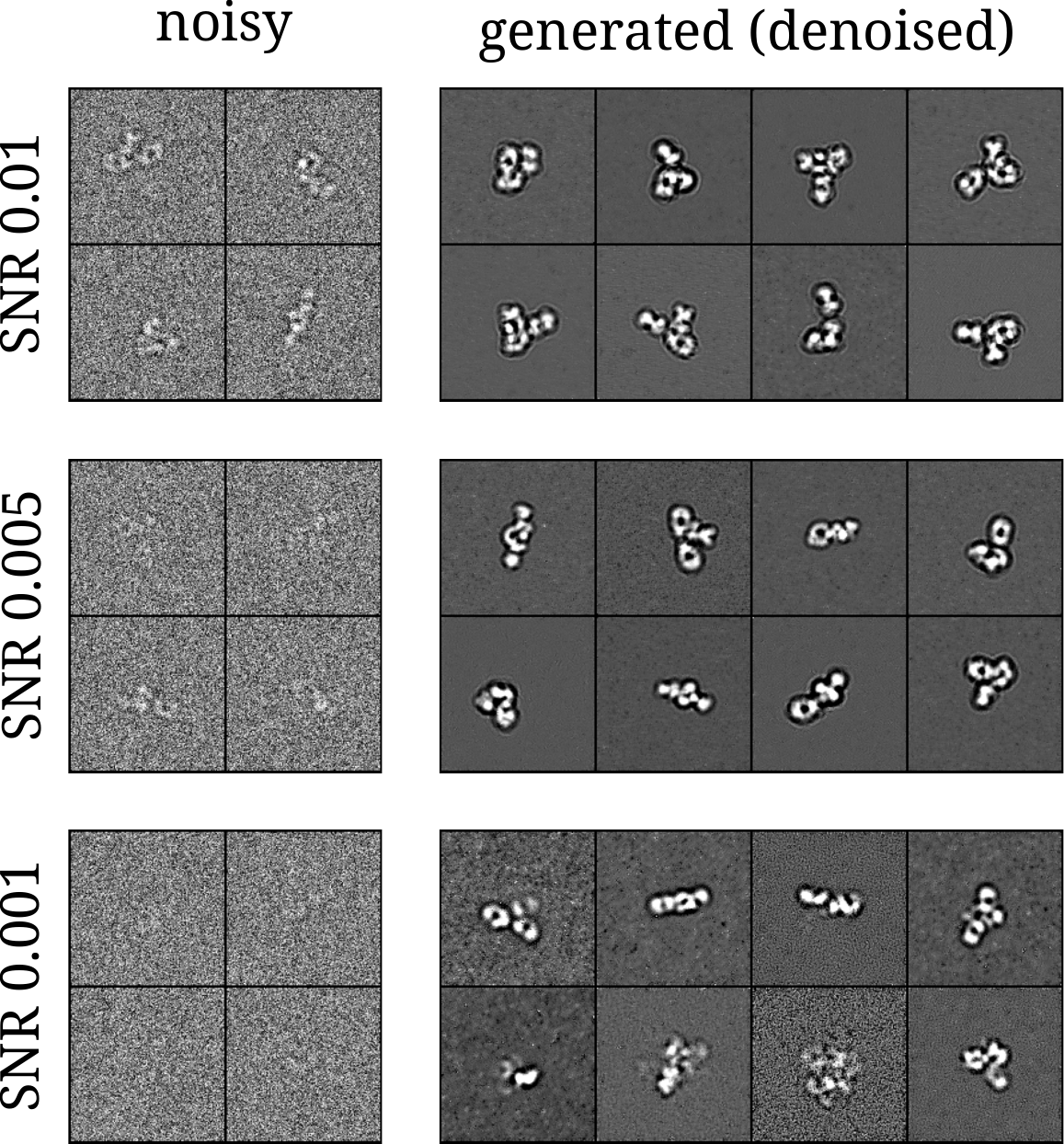}
\caption{NR-CFM samples on the CryoBench IgG-1D dataset. Samples are shown for three SNR levels.
Bold denotes the best result.
The left panels show noisy training observations, and the right panels show samples generated by NR-CFM.
}
\label{fig:cryoet_benchmark}
\end{figure}

To test NR-CFM beyond natural images, we use the CryoBench IgG-1D cryo-EM
dataset \cite{jeon2024cryobench} as a severe-noise generation stress test,
rather than for 3D reconstruction. As shown in
Figure~\ref{fig:cryoet_benchmark}, NR-CFM produces remarkably clean particle
images even in this extreme-noise setting.

\begin{table}[t]
\centering
\caption{FID ($\downarrow$) on CryoBench IgG-1D. Values are mean $\pm$ standard deviation.}
\label{tab:cfm_snr_comparison}
\scriptsize
\resizebox{\linewidth}{!}{%
\begin{tabular}{lccc}
\toprule
method & snr $0.01$ & snr $0.005$ & snr $0.001$ \\
\midrule
CFM & $422.15 \pm 0.19$ & $424.42 \pm 0.13$ & $431.18 \pm 0.12$ \\
NR-CFM (ours) & \textbf{157.80 $\pm$ 0.05} & \textbf{180.44 $\pm$ 0.14} & \textbf{223.97 $\pm$ 0.07} \\
\bottomrule
\end{tabular}}
\end{table}
\section*{Discussion}
\label{sec:discussion}
NR-CFM is competitive when the corruption process can be modeled explicitly,
with encouraging results on structured Gaussian noise. Under AWGN, the gap to
Ambient Diffusion narrows as noise increases. At $\sigma=0.2$, NR-CFM achieves
lower FID than the reported Ambient Diffusion results by 2.22 points on CIFAR-10
and 2.88 points on CelebA-HQ. This suggests that the correction is particularly
useful under severe noise. The low-corruption gap
may partly reflect model maturity: Ambient Diffusion uses the highly optimized
EDM backbone, whereas CFM is newer. This is supported by the gap on clean data
at $\sigma=0$, where corruption correction is unnecessary and the comparison
primarily reflects the underlying generative models. More optimized CFM
architectures and training recipes may therefore further narrow the gap.

On the CryoBench IgG-1D stress test, NR-CFM generates recognizable particle
structure even at SNR $0.001$. Although qualitative and not a full cryo-EM
reconstruction benchmark, this result suggests that noise-robust flow matching
may be useful in low-SNR scientific imaging beyond natural images.

The main limitations of the current framework are its reliance on a known
corruption model, limited flexibility across noise types, and its current
representational scope. Requiring access to the corruption mechanism is a
meaningful restriction relative to NR-GAN, which is designed to operate without
it, although this assumption may be less severe in domains where the
measurement process is known or can be modeled accurately, such as medical
imaging; when the noise model is misspecified, poorly characterized, or changes
over time, practical robustness may degrade substantially. In addition, each
trained model currently targets only a single noise type, so although the
framework covers a broad class of Gaussian corruptions, separate corruption
families or parameter settings still require separate models, making deployment
harder when datasets contain multiple corruption sources or when corruption
changes between training and inference. The corruption process must fit the
Gaussian model with sample-specific covariance. This covers many structured and
signal-dependent noise patterns. Some count-based corruptions can also be
approximated by a Gaussian model at high count levels. Extending NR-CFM beyond
this Gaussian model remains an important direction for future work.

These limitations suggest several directions for future work: relaxing the
known-noise assumption, learning a single model that handles a family of
corruptions, and extending the framework beyond the general Gaussian model.
Progress on these questions would make the method more useful when corruption
mechanisms are only partially known.
\section*{Conclusion}
We studied generative modeling from corrupted-only observations in the conditional flow matching setting and introduced Noise-Robust Conditional Flow Matching (NR-CFM), a framework for recovering clean samples from CFM models trained without clean data. The method learns the flow of an observable noisy bridge and then corrects that flow toward the clean distribution, yielding a closed-form solution in AWGN and an extension to structured and signal-dependent Gaussian corruption through a learned general Gaussian correction. Empirically, NR-CFM improves over NR-GAN on the corruption settings covered by our formulation and remains competitive with ambient diffusion in the known-AWGN regime. These results show that conditional flow matching can serve as a strong foundation for noise-robust generation from corrupted-only observations, while suggesting high-noise scientific imaging as a promising direction for future study.
\section*{Acknowledgments}
This work was partially funded by the Center for Advanced Systems Understanding (CASUS) which is financed by Germany’s Federal Ministry of Research, Technology and Space (BMFTR) and by the Saxon Ministry for Science, Culture, and Tourism (SMWK) with tax funds based on the budget approved by the Saxon State Parliament. AU and AY were supported by the Helmholtz Association Initiative and Networking Fund in the frame of Helmholtz AI within the project “tomoCAT". Furthermore, AY was suported by by the Helmholtz Foundation Model Initiative within the project “PROFOUND”. The authors gratefully acknowledge the Gauss Centre for Supercomputing e.V. (www.gauss-centre.eu) for funding this project by providing computing time through the John von Neumann Institute for Computing (NIC) on the GCS Supercomputer JUWELS at Jülich Supercomputing Centre (JSC).

\bibliographystyle{plainnat}
\bibliography{references}

@inproceedings{lipman2023flow,
  title={Flow Matching for Generative Modeling},
  author={Lipman, Yaron and Chen, Ricky TQ and Ben-Hamu, Heli and Nickel, Maximilian and Le, Matt},
  booktitle={11th International Conference on Learning Representations, ICLR 2023},
  year={2023}
}

@inproceedings{liu2023flow,
  title={Flow Straight and Fast: Learning to Generate and Transfer Data with Rectified Flow},
  author={Liu, Xingchao and Gong, Chengyue and Liu, Qiang},
  booktitle={The Eleventh International Conference on Learning Representations (ICLR)},
  year={2023}
}

@inproceedings{albergo2023building,
  title={Building Normalizing Flows with Stochastic Interpolants},
  author={Albergo, Michael and Vanden-Eijnden, Eric},
  booktitle={ICLR 2023 Conference},
  year={2023}
}

@inproceedings{sohl2015deep,
  title={Deep unsupervised learning using nonequilibrium thermodynamics},
  author={Sohl-Dickstein, Jascha and Weiss, Eric and Maheswaranathan, Niru and Ganguli, Surya},
  booktitle={International conference on machine learning},
  pages={2256--2265},
  year={2015},
  organization={pmlr}
}

@article{ho2020denoising,
  title={Denoising diffusion probabilistic models},
  author={Ho, Jonathan and Jain, Ajay and Abbeel, Pieter},
  journal={Advances in neural information processing systems},
  volume={33},
  pages={6840--6851},
  year={2020}
}

@inproceedings{song2021score,
  title={Score-Based Generative Modeling through Stochastic Differential Equations},
  author={Song, Yang and Sohl-Dickstein, Jascha and Kingma, Diederik P and Kumar, Abhishek and Ermon, Stefano and Poole, Ben},
  booktitle={International Conference on Learning Representations},
  year={2021}
}

@inproceedings{jeon2024cryobench,
  title={CryoBench: Diverse and challenging datasets for the heterogeneity problem in cryo-EM},
  author={Jeon, Minkyu and Raghu, Rishwanth and Astore, Miro and Woollard, Geoffrey and Feathers, Ryan and Kaz, Alkin and Hanson, Sonya M. and Cossio, Pilar and Zhong, Ellen D.},
  booktitle={Advances in Neural Information Processing Systems},
  year={2024}
}

@inproceedings{kaneko2020noise,
  title={Noise robust generative adversarial networks},
  author={Kaneko, Takuhiro and Harada, Tatsuya},
  booktitle={Proceedings of the IEEE/CVF conference on computer vision and pattern recognition},
  pages={8404--8414},
  year={2020}
}

@inproceedings{bora2018ambientgan,
  title={AmbientGAN: Generative models from lossy measurements},
  author={Bora, Ashish and Price, Eric and Dimakis, Alexandros G},
  booktitle={International conference on learning representations},
  year={2018}
}

@article{daras2024ambientlaws,
  title={Ambient diffusion: Learning clean distributions from corrupted data},
  author={Daras, Giannis and Shah, Kulin and Dagan, Yuval and Gollakota, Aravind and Dimakis, Alex and Klivans, Adam},
  journal={Advances in Neural Information Processing Systems},
  volume={36},
  pages={288--313},
  year={2023}
}

@article{daras2024much,
  title={How much is a noisy image worth? data scaling laws for ambient diffusion},
  author={Daras, Giannis and Cherapanamjeri, Yeshwanth and Daskalakis, Constantinos},
  journal={arXiv preprint arXiv:2411.02780},
  year={2024}
}

@inproceedings{daras2024consistent,
  title={Consistent Diffusion Meets Tweedie: Training Exact Ambient Diffusion Models with Noisy Data},
  author={Daras, Giannis and Dimakis, Alex and Daskalakis, Constantinos Costis},
  booktitle={Forty-first International Conference on Machine Learning},
  year={2024}
}

@article{bai2024expectation,
  title={An expectation-maximization algorithm for training clean diffusion models from corrupted observations},
  author={Bai, Weimin and Wang, Yifei and Chen, Wenzheng and Sun, He},
  journal={Advances in Neural Information Processing Systems},
  volume={37},
  pages={19447--19471},
  year={2024}
}

@inproceedings{lu2025stochastic,
  title={Stochastic Forward--Backward Deconvolution: Training Diffusion Models with Finite Noisy Datasets},
  author={Lu, Haoye and Wu, Qifan and Yu, Yaoliang},
  booktitle={International Conference on Machine Learning},
  pages={40741--40768},
  year={2025},
  organization={PMLR}
}

@article{krizhevsky2009learning,
  title={Learning multiple layers of features from tiny images},
  author={Krizhevsky, Alex and Hinton, Geoffrey and others},
  year={2009},
  publisher={Toronto, ON, Canada}
}

@inproceedings{liu2015deep,
  title={Deep learning face attributes in the wild},
  author={Liu, Ziwei and Luo, Ping and Wang, Xiaogang and Tang, Xiaoou},
  booktitle={Proceedings of the IEEE international conference on computer vision},
  pages={3730--3738},
  year={2015}
}

@article{yu2015lsun,
  title={Lsun: Construction of a large-scale image dataset using deep learning with humans in the loop},
  author={Yu, Fisher and Seff, Ari and Zhang, Yinda and Song, Shuran and Funkhouser, Thomas and Xiao, Jianxiong},
  journal={arXiv preprint arXiv:1506.03365},
  year={2015}
}

@inproceedings{karras2019style,
  title={A style-based generator architecture for generative adversarial networks},
  author={Karras, Tero and Laine, Samuli and Aila, Timo},
  booktitle={Proceedings of the IEEE/CVF conference on computer vision and pattern recognition},
  pages={4401--4410},
  year={2019}
}

@article{kawar2024gsure,
  author        = {Kawar, Bahjat and Elata, Noam and
                   Michaeli, Tomer and Elad, Michael},
  title         = {{GSURE}-Based Diffusion Model Training with Corrupted Data},
  journal       = {Transactions on Machine Learning Research},
  year          = {2024},
  eprint        = {2305.13128},
  archivePrefix = {arXiv},
  primaryClass  = {cs.LG}
}

@article{kelkar2024ambientflow,
  author        = {Kelkar, Varun A. and Deshpande, Rucha and
                   Banerjee, Arindam and Anastasio, Mark A.},
  title         = {{AmbientFlow}: Invertible generative models from incomplete,
                   noisy measurements},
  journal       = {Transactions on Machine Learning Research},
  year          = {2024},
  eprint        = {2309.04856},
  archivePrefix = {arXiv},
  primaryClass  = {cs.LG}
}

@inproceedings{zhang2025inverse,
  author    = {Zhang, Yuchen and Zhou, Jian},
  title     = {Inverse Flow and Consistency Models},
  booktitle = {Proceedings of the 42nd International Conference on Machine Learning},
  series    = {Proceedings of Machine Learning Research},
  volume    = {267},
  pages     = {77152--77176},
  year      = {2025},
  publisher = {PMLR},
  eprint    = {2502.11333},
  archivePrefix = {arXiv},
  primaryClass  = {cs.LG}
}

@inproceedings{modi2026scsi,
  author    = {Modi, Chirag and Han, Jiequn and
               Vanden-Eijnden, Eric and Bruna, Joan},
  title     = {Generative Modeling from Black-Box Corruptions via
               Self-Consistent Stochastic Interpolants},
  booktitle = {International Conference on Learning Representations},
  year      = {2026},
  eprint    = {2512.10857},
  archivePrefix = {arXiv},
  primaryClass  = {cs.LG}
}

@inproceedings{lehtinen2018noise2noise,
  author    = {Lehtinen, Jaakko and Munkberg, Jacob and Hasselgren, Jon
               and Laine, Samuli and Karras, Tero and Aittala, Miika
               and Aila, Timo},
  title     = {{Noise2Noise}: Learning Image Restoration without Clean Data},
  booktitle = {Proceedings of the 35th International Conference on Machine Learning},
  series    = {Proceedings of Machine Learning Research},
  volume    = {80},
  pages     = {2965--2974},
  year      = {2018},
  publisher = {PMLR},
  url       = {https://proceedings.mlr.press/v80/lehtinen18a.html}
}

@inproceedings{krull2019noise2void,
  author    = {Krull, Alexander and Buchholz, Tim-Oliver and Jug, Florian},
  title     = {{Noise2Void} - Learning Denoising From Single Noisy Images},
  booktitle = {Proceedings of the IEEE/CVF Conference on Computer Vision and Pattern Recognition},
  pages     = {2129--2137},
  year      = {2019},
  month     = jun,
  doi       = {10.1109/CVPR.2019.00223},
  url       = {https://openaccess.thecvf.com/content_CVPR_2019/html/Krull_Noise2Void_-_Learning_Denoising_From_Single_Noisy_Images_CVPR_2019_paper.html}
}

@inproceedings{batson2019noise2self,
  author    = {Batson, Joshua and Royer, Loic},
  title     = {{Noise2Self}: Blind Denoising by Self-Supervision},
  booktitle = {Proceedings of the 36th International Conference on Machine Learning},
  series    = {Proceedings of Machine Learning Research},
  volume    = {97},
  pages     = {524--533},
  year      = {2019},
  publisher = {PMLR},
  url       = {https://proceedings.mlr.press/v97/batson19a.html}
}

@inproceedings{soltanayev2018sure,
  author    = {Soltanayev, Shakarim and Chun, Se Young},
  title     = {Training Deep Learning Based Denoisers without Ground Truth Data},
  booktitle = {Advances in Neural Information Processing Systems},
  volume    = {31},
  pages     = {3257--3267},
  year      = {2018},
  publisher = {Curran Associates, Inc.},
  eprint    = {1803.01314},
  archivePrefix = {arXiv},
  primaryClass  = {cs.CV},
  url       = {https://proceedings.neurips.cc/paper_files/paper/2018/hash/c0560792e4a3c79e62f76cbf9fb277dd-Abstract.html}
}

@inproceedings{moran2020noisier2noise,
  author    = {Moran, Nick and Schmidt, Dan and Zhong, Yu and Coady, Patrick},
  title     = {{Noisier2Noise}: Learning to Denoise From Unpaired Noisy Data},
  booktitle = {Proceedings of the IEEE/CVF Conference on Computer Vision and
               Pattern Recognition},
  pages     = {12064--12072},
  year      = {2020},
  month     = jun,
  doi       = {10.1109/CVPR42600.2020.01208}
}

@inproceedings{pang2021recorrupted,
  author    = {Pang, Tongyao and Zheng, Huan and Quan, Yuhui and Ji, Hui},
  title     = {Recorrupted-to-Recorrupted: Unsupervised Deep Learning for
               Image Denoising},
  booktitle = {Proceedings of the IEEE/CVF Conference on Computer Vision and
               Pattern Recognition},
  pages     = {2043--2052},
  year      = {2021},
  month     = jun,
  doi       = {10.1109/CVPR46437.2021.00208}
}

@inproceedings{kim2021noise2score,
  author    = {Kim, Kwanyoung and Ye, Jong Chul},
  title     = {{Noise2Score}: {Tweedie's} Approach to Self-Supervised Image
               Denoising without Clean Images},
  booktitle = {Advances in Neural Information Processing Systems},
  volume    = {34},
  pages     = {864--874},
  year      = {2021},
  publisher = {Curran Associates, Inc.},
  url       = {https://proceedings.neurips.cc/paper_files/paper/2021/hash/077b83af57538aa183971a2fe0971ec1-Abstract.html}
}

@inproceedings{aali2023surescore,
  author    = {Aali, Asad and Arvinte, Marius and Kumar, Sidharth and
               Tamir, Jonathan I.},
  title     = {Solving Inverse Problems with Score-Based Generative Priors
               Learned from Noisy Data},
  booktitle = {2023 57th Asilomar Conference on Signals, Systems, and
               Computers},
  pages     = {837--843},
  year      = {2023},
  publisher = {IEEE},
  doi       = {10.1109/IEEECONF59524.2023.10477042}
}

@inproceedings{rozet2024learning,
  author    = {Rozet, Fran{\c{c}}ois and Andry, G{\'e}r{\^o}me and
               Lanusse, Fran{\c{c}}ois and Louppe, Gilles},
  title     = {Learning Diffusion Priors from Observations by
               Expectation Maximization},
  booktitle = {Advances in Neural Information Processing Systems},
  volume    = {37},
  pages     = {87647--87682},
  year      = {2024},
  publisher = {Curran Associates, Inc.},
  doi       = {10.52202/079017-2783}
}

\onecolumn
\appendix

\onecolumn
\appendix
\setcounter{secnumdepth}{2}
\renewcommand{\thesubsection}{\thesection.\arabic{subsection}}

\section{Implementation and Evaluation Details}
\label{app:implementation_details}

This section collects the dataset, corruption, evaluation, training, and
sampling details for the reported experiments.

\subsection{NR-GAN Comparison}
\label{app:nrgan_details}

\paragraph{Evaluation and datasets.}
Following NR-GAN \cite{kaneko2020noise}, we evaluate Fr\'echet Inception
Distance (FID) using $10{,}000$ generated samples and test-set reference
statistics.  For CIFAR-10, we use $50{,}000$ training and $10{,}000$ test
images at $32\times32$ resolution.  LSUN Bedroom and FFHQ use the dataset
splits and image resolutions reported by NR-GAN.  Images are normalized to
$[-1,1]$ throughout.

\paragraph{Corruption settings.}
NR-GAN defines sixteen settings (A)--(P). We evaluate the Gaussian settings
(A)--(D) and (G)--(L) on CIFAR-10, A, B, G, I, and L on LSUN Bedroom, and A and
I on FFHQ. The signal-independent settings are: (A) AGF, additive Gaussian
noise with fixed $\sigma=0.2$; (B) AGV, additive Gaussian noise with per-image
$\sigma\in[0.04,0.4]$; (C) LGF, Gaussian noise with $\sigma=0.2$ on a fixed
$16\times16$ patch; (D) LGV, the same noise on a patch sampled from
$[8,24]\times[8,24]$; (G) BG, spatially correlated brown Gaussian noise; and
(H) A+G, white plus brown Gaussian noise. The signal-dependent settings are:
(I) MGF, multiplicative Gaussian noise with fixed $\sigma=0.2$; (J) MGV, the
same with per-image $\sigma\in[0.04,0.4]$; and (K,L) A+I, additive plus
multiplicative Gaussian noise with respective
$(\sigma_{\mathrm{add}},\sigma_{\mathrm{mult}})$ values $(0.04,0.2)$ and
$(0.2,0.2)$. These magnitudes use the normalized $[-1,1]$ scale rather than the
original $[0,255]$ convention of \cite{kaneko2020noise}.

\paragraph{Training.}
The stage-1 flow-matching model uses a DDPM++ architecture with base width
$128$, channel multipliers $(1,2,2,2)$, four residual blocks per level,
dropout $0$, attention at resolution $16$, single-head output, and gradient
checkpointing enabled.  Training uses $4$ A100 GPUs with per-GPU batch size
$128$ for CIFAR-10 and $64$ for LSUN Bedroom and FFHQ.  We use gradient
clipping at norm $1.0$, EMA decay $0.9999$, a $5000$-step warmup, and a
$200{,}000$-step training budget.  The flow-matching objective uses $\ell_2$
loss, weighting enabled with $k=1.0$, and endpoint regularization
$\varepsilon=10^{-3}$.

In stage 2, the stage-1 primary network is frozen and reused.  The correction
network retains the same base width, channel multipliers, dropout, and
attention resolution, but uses two residual blocks per level and a two-head
output.  With the same hardware and batch sizes, we use gradient clipping at
norm $1.0$, EMA decay $0.9999$, a $2000$-step warmup, and a $200{,}000$-step
training budget.  The auxiliary objective uses weight $1.0$ and
$\delta\in[0.01,0.3]$ drawn from a log-uniform schedule.

\paragraph{Sampling.}
For these experiments, we use a conservative sampling cutoff of
$t_{\mathrm{cut}}=0.95$.  Across the broad range of evaluated settings, the
modest performance cost of stopping early is preferable to the sharp
degradation that can occur when the cutoff is placed too close to the endpoint.
We use the adaptive Dormand--Prince (dopri5) solver with absolute and relative
tolerances of $10^{-5}$.

\subsection{Ambient Diffusion Comparison}
\label{app:ambient_diffusion_details}

\paragraph{Evaluation and datasets.}
Following \cite{daras2024much}, we evaluate FID using $50{,}000$ generated
samples, three random seeds, and reference statistics from the full image set
used by each model.  We use CIFAR-10 and CelebA-HQ in the setup of
\cite{daras2024ambientlaws,daras2024much}, with AWGN standard deviations
$\sigma\in\{0,0.05,0.1,0.2\}$.  Images are normalized to $[-1,1]$.
In this protocol, both the generated distribution and the reference
distribution are represented by $50{,}000$ samples when computing FID.  This
larger, balanced sample count reduces finite-sample estimation bias and
therefore produces lower FID values than the NR-GAN protocol above, which uses
only $10{,}000$ generated and reference samples.  Consequently, the two CIFAR-10 comparisons
are not directly equivalent.

\paragraph{Training and sampling.}
We use the ADM architecture of Dhariwal and Nichol
for CIFAR-10 and CelebA-HQ, following the
ImageNet-$64$ recipe of Lipman et al.~\cite{lipman2023flow}.  Training uses
AdamW with zero weight decay, gradient-norm clipping at $1.0$, dropout $0.1$,
EMA decay $0.9999$, and $400{,}000$ steps on $4$ A100 GPUs.  The primary model
is trained with the plain flow-matching MSE objective, without optimal-transport
coupling or loss reweighting.  The per-GPU batch size is $64$ for CIFAR-10
(effective batch size $256$) and $128$ for CelebA-HQ (effective batch size
$512$).

For CIFAR-10, we use a peak learning rate of $2\times10^{-4}$ with a
$5000$-step warmup followed by cosine decay, and train in full $32$-bit
precision.  For CelebA-HQ, we use a constant learning rate of $10^{-4}$ and
bfloat16 mixed precision.  The backbones are capacity-matched to the EDM
DDPM++ baselines rather than reproducing their architectures exactly.  The
CIFAR-10 model has approximately $55$M parameters, with $128$ base channels,
channel multipliers $(1,2,2,2)$, three residual blocks per level, and attention
at $16\times16$ resolution.  The CelebA-HQ model has approximately $60$M
parameters and uses the same backbone with an additional attention resolution
at $8\times8$.  These sizes closely match the approximately $55.7$M and $60$M
EDM DDPM++ models used by Daras et al.~\cite{daras2024much}, so the NR-CFM and
Ambient Diffusion comparison is not confounded by model capacity.

Sampling uses $t_{\mathrm{cut}}=0.98$, which provides a robust compromise
across the noise levels in Table~\ref{tab:t_cutoff_ablation}.  We use the
adaptive Dormand--Prince (dopri5) solver with absolute and relative tolerances
of $10^{-5}$.

\subsection{CryoBench Experiment}
\label{app:cryobench_details}

\paragraph{Dataset and image formation.}
The IgG-1D subset of CryoBench is a controlled conformational-heterogeneity
dataset in which one domain of an immunoglobulin G (IgG) antibody complex is
rotated through a full $360^\circ$ cycle, producing a one-dimensional circular
trajectory of structures. The released dataset contains 100 ground-truth atomic
models and corresponding volumes, together with simulated cryo-EM image stacks
at multiple signal-to-noise ratios. CryoBench generates these synthetic
particles by converting each atomic model into a 3D electron-scattering density
volume and simulating the standard cryo-EM forward model. In the notation of
\cite{jeon2024cryobench}, each image is formed in the Fourier domain as
\[
I_i = C_i P_{\phi_i} V_i + \eta_i,
\]
where $V_i$ is the density volume for the sampled conformation,
$P_{\phi_i}$ is the projection operator for pose
$\phi_i=(R_i,t_i)\in SO(3)\times\mathbb{R}^2$, $C_i$ is the contrast transfer
function (CTF), and $\eta_i$ is additive white Gaussian noise (AWGN). Thus,
after sampling a conformation along the IgG-1D trajectory, CryoBench samples a
viewing pose, projects the corresponding 3D volume to a 2D particle image,
applies the CTF in Fourier space, and corrupts the result with AWGN at the
specified signal-to-noise ratio (SNR).

For each SNR level, we estimate the observation-noise mean and standard
deviation from the four $16\times16$ corner patches of every image.  We pool
all pixels from these patches and compute their empirical mean and standard
deviation.  Because the image corners are assumed to contain no projected
molecular structure, these pixels provide samples of the background noise.
We then subtract the estimated mean from every image and divide by the
estimated standard deviation, so that the normalized background noise is
approximately zero-mean with unit standard deviation.

\paragraph{Architecture, training, and sampling.}
For CryoBench, we use a DDPM++ U-Net at $128\times128$ resolution with $128$
base channels, channel multipliers $(1,2,2,2)$, two residual blocks per level,
attention at $16\times16$ resolution, and approximately $36$M parameters.  We
train with AdamW using weight decay $0.01$, gradient-norm clipping at $1.0$,
and EMA decay $0.9999$.  The learning rate is $3\times10^{-4}$ after a
$5000$-step warmup and is then cosine-annealed to $10^{-4}$ over a total of
$300{,}000$ steps.  Training uses full $32$-bit precision on $4$ A100 GPUs.

Unlike the CIFAR-10 and CelebA-HQ experiments, we apply no additional synthetic
corruption: the raw CryoBench particle images serve directly as the noisy
observations, without paired clean targets.  The background-based normalization
described above is consistent with the $\mathcal{N}(0,I)$ prior scale.  At
sampling time, we use the same closed-form
AWGN correction $\widehat{s}=\sigma^2 g_\theta$ as for CIFAR-10 and CelebA-HQ,
with $\sigma\approx1$ in the globally normalized space.

For evaluation, we compute FID over three independently generated sets of
$50{,}000$ images, using seed ranges $0$--$49{,}999$, $50{,}000$--$99{,}999$,
and $100{,}000$--$149{,}999$.  Reference statistics are computed from clean
IgG-1D particle renderings produced with the published CryoBench simulator and
provided particle poses.  Sampling uses the adaptive dopri5 solver with
absolute and relative tolerances of $10^{-5}$ and cutoff
$t_{\mathrm{cut}}=0.95$.

\paragraph{Clean reference generation.}
Because CryoBench releases only $D{=}128$ volumes, we first zero-pad each
conformation's volume in the Fourier domain to $D{=}256$ to recover the native
simulation resolution ($\mathrm{Apix}=1.5$) at which the CTF was originally
applied. We then reuse the released ground-truth poses $\phi_i$ and CTF
parameters $C_i$ for each image, applying the identical forward model
$I_i = C_i P_{\phi_i} V_i$ but omitting the noise term ($\eta_i=0$), so that
each clean reference image shares its conformation, pose, and CTF exactly with
one released noisy image. The resulting $256\times256$ images are Fourier-
cropped back to $D{=}128$, matching the downsampling convention used to
produce the released noisy stacks, and the contrast sign is matched to the
released convention. This yields a paired noiseless counterpart to the
CryoBench IgG-1D dataset, which we use both to compute FID reference
statistics and, at matched sample sizes, to establish the FID floor induced by
the domain gap between cryo-EM particle images and the ImageNet-pretrained
Inception features.

\section{Notation and Proofs}
\label{app:proof_preliminaries}

We use uppercase letters for random variables and lowercase letters for their
realizations.  The clean endpoint, prior endpoint, and observed endpoint are
\[
    X_1\sim p_{\mathrm{data}},
    \qquad
    X_0\sim\mathcal{N}(0,I),
    \qquad
    Y=X_1+\varepsilon_y.
\]
The prior endpoint \(X_0\) is independent of
\((X_1,\Omega,\varepsilon_y)\).  The clean and observed bridges are
\[
    X_t=(1-t)X_0+tX_1,
    \qquad
    Y_t=(1-t)X_0+tY
    =
    X_t+t\varepsilon_y.
\]
We write \(p_t^{\mathrm{clean}}\) and \(p_t^{\mathrm{noisy}}\) for the densities
of \(X_t\) and \(Y_t\), respectively.  In the appendix, \(y\) denotes a
realization of \(Y_t\) at a fixed time \(t\).  The same realized state is
written as \(y_t\) in the main text.

The population velocity of the observed bridge is
\[
    v_t^\star(y)
    :=
    \mathbb{E}[Y-X_0\mid Y_t=y,t].
\]
We also use the posterior mean observation noise
\[
    e_t(y)
    :=
    \mathbb{E}[\varepsilon_y\mid Y_t=y,t].
\]
The corresponding correction field is
\[
    s_t(y)
    :=
    -\frac{1}{t}e_t(y),
    \qquad t>0.
\]
To match the notation in the Methods section, we define the
population score anchor by
\[
    g_t^\star(y)
    :=
    \frac{t\,v_t^\star(y)-y}{1-t},
    \qquad t<1.
\]
Appendix~\ref{app:proof_score_awgn} shows that \(g_t^\star(y)\) equals
\(\nabla_y\log p_t^{\mathrm{noisy}}(y)\) for \(t\in(0,1)\).  The correction
\(s_t\) is defined for \(t>0\), while the score anchor is used only for \(t<1\).
Endpoint expressions are interpreted through limits when those limits are
explicitly required.

For the general Gaussian corruption model, let
\[
    S:=\Sigma(X_1,\Omega)
\]
denote the noise covariance for the sample. For a fixed clean sample and noise
setting, the observation noise is Gaussian with covariance $S$:
\[
    \varepsilon_y\mid X_1,\Omega
    \sim
    \mathcal{N}(0,S).
\]
Before conditioning, $S$ varies across samples; after fixing $(X_1,\Omega)$,
the covariance is fixed for that sample.

For the derivations below, we assume that \(S\) is almost surely symmetric
positive definite.  Singular or low-rank corruptions can instead be handled by
restricting the derivation to the support of \(S\), or by using the regularized
covariance \(S+\eta I\) and taking \(\eta\rightarrow0\).

For a matrix field
\(M:\mathbb{R}^d\rightarrow\mathbb{R}^{d\times d}\), we use the row-wise
divergence convention
\begin{equation}
    \bigl(\operatorname{div}M(y)\bigr)_i
    =
    \sum_{j=1}^d
    \partial_{y_j}M_{ij}(y).
    \label{eq:app_divergence_convention}
\end{equation}
With this convention,
\begin{equation}
    \operatorname{div}\!\bigl(p(y)M(y)\bigr)
    =
    M(y)\nabla_y p(y)
    +
    p(y)\operatorname{div}M(y).
    \label{eq:app_matrix_product_rule}
\end{equation}
Throughout the appendix, we assume that the densities, posterior expectations,
and spatial derivatives required by each derivation exist.  We also assume that
differentiation can be exchanged with integration and that boundary terms vanish
when integration by parts is used.  The same identities may be interpreted
distributionally under standard approximation arguments.

\subsection{Gaussian Tweedie identity.}

We repeatedly use the following Gaussian Tweedie identity.
Let
\[
    B=A+\xi,
    \qquad
    \xi\sim\mathcal{N}(0,\Gamma),
\]
where \(\xi\) is independent of \(A\) and \(\Gamma\succ0\).  If \(p_B\) is
differentiable, then
\begin{equation}
    \mathbb{E}[A\mid B=b]
    =
    b+\Gamma\nabla_b\log p_B(b).
    \label{eq:app_tweedie_identity}
\end{equation}

\paragraph{Proof.}
Let
\[
    q(b\mid a)=\mathcal{N}(b;a,\Gamma)
\]
denote the conditional density of \(B\) given \(A=a\).  The marginal density is
\[
    p_B(b)
    =
    \int q(b\mid a)\,dP_A(a).
\]
Differentiating with respect to \(b\) gives
\[
    \nabla_b p_B(b)
    =
    -\Gamma^{-1}
    \int
    (b-a)q(b\mid a)\,dP_A(a).
\]
Multiplying by \(\Gamma\),
\begin{equation}
\begin{aligned}
    \Gamma\nabla_b p_B(b)
    &=
    \int
    (a-b)q(b\mid a)\,dP_A(a)
    \\
    &=
    p_B(b)
    \left(
        \mathbb{E}[A\mid B=b]-b
    \right).
\end{aligned}
\end{equation}
Dividing by \(p_B(b)\) proves \eqref{eq:app_tweedie_identity}.
\hfill\(\square\)

\subsection{Observed Flow Matching}
\label{app:proof_observed_flow}

Conditioned on the observed endpoint \(Y=y\), the observed bridge is
\[
    Y_t=(1-t)X_0+ty.
\]
Since \(X_0\sim\mathcal{N}(0,I)\),
\begin{equation}
    Y_t\mid Y=y
    \sim
    \mathcal{N}\!\left(ty,(1-t)^2I\right).
    \label{eq:app_observed_conditional_path}
\end{equation}
The samplewise velocity along this conditional path is
\[
    \frac{d}{dt}\bigl((1-t)X_0+tY\bigr)=Y-X_0.
\]
Therefore the observed CFM objective
\[
    \mathcal{L}_{\mathrm{obs}}(\theta)
    =
    \mathbb{E}
    \left[
        \left\|
        v_\theta(Y_t,t)-(Y-X_0)
        \right\|^2
    \right]
\]
is a least-squares regression problem.  For fixed \((Y_t,t)=(y,t)\), the
population minimizer is the conditional mean of the target:
\begin{equation}
    v_t^\star(y)
    =
    \mathbb{E}[Y-X_0\mid Y_t=y,t].
    \label{eq:app_observed_velocity}
\end{equation}
This establishes that \(v_t^\star\) is the velocity field of the observed
bridge, whose endpoint law is the law of \(Y\).  It does not by itself provide a
clean-data velocity field.

\subsection{AWGN Path Dynamics}
\label{app:proof_path_dynamics}

Assume in this subsection that
\[
    \varepsilon_y\sim\mathcal{N}(0,\sigma^2I)
\]
is independent of \((X_0,X_1)\).  Conditioning on \(X_1=x_1\), the clean bridge
is
\[
    X_t=(1-t)X_0+tx_1,
    \qquad
    X_0\sim\mathcal{N}(0,I).
\]
Therefore
\begin{equation}
    X_t\mid X_1=x_1
    \sim
    \mathcal{N}\!\left(tx_1,(1-t)^2I\right).
    \label{eq:app_clean_path_conditional}
\end{equation}
Since
\[
    Y_t=X_t+t\varepsilon_y
\]
and \(t\varepsilon_y\sim\mathcal{N}(0,\sigma^2t^2I)\) is independent of
\(X_t\), the conditional noisy bridge is
\begin{equation}
    Y_t\mid X_1=x_1
    \sim
    \mathcal{N}
    \left(
        tx_1,
        \bigl((1-t)^2+\sigma^2t^2\bigr)I
    \right).
    \label{eq:app_awgn_conditional_noisy_path}
\end{equation}
Equivalently, if \(G_t=\mathcal{N}(0,\sigma^2t^2I)\), then
\begin{equation}
    p_t^{\mathrm{noisy}}(y)
    =
    \int p_t^{\mathrm{clean}}(x)G_t(y-x)\,dx
    =
    \bigl(p_t^{\mathrm{clean}}*G_t\bigr)(y).
    \label{eq:app_awgn_noisy_marginal}
\end{equation}
At \(t=0\), \(G_t\) degenerates to a Dirac mass and the identity is understood
as a weak limit.

We now derive the transport--diffusion form of this marginal evolution.  The
original bridge
\[
    Y_t=X_t+t\varepsilon_y
\]
uses a single noise realization \(\varepsilon_y\) that is fixed across time.
Thus it is not literally a Brownian diffusion path in \(t\).  However, for the
purpose of deriving one-time marginal densities, it is equivalent to replacing
\(t\varepsilon_y\) by a Brownian perturbation with the same covariance.  Let
\((B_t)_{t\ge0}\) be a standard Brownian motion independent of
\((X_0,X_1)\), and define
\begin{equation}
    R_t
    :=
    \int_0^t \sqrt{2\sigma^2s}\,dB_s.
    \label{eq:app_awgn_brownian_noise}
\end{equation}
Then
\[
    R_t\sim\mathcal{N}(0,\sigma^2t^2I),
    \qquad
    \operatorname{Cov}(R_t)
    =
    \int_0^t 2\sigma^2s\,ds\,I
    =
    \sigma^2t^2I.
\]
Therefore
\[
    \widetilde{Y}_t:=X_t+R_t
\]
has the same marginal law as \(Y_t=X_t+t\varepsilon_y\) at each fixed time
\(t\).  Moreover, because both \(R_t\) and \(t\varepsilon_y\) are independent
Gaussian perturbations of \(X_t\) with covariance \(\sigma^2t^2I\), the joint
law of
\[
    (X_t,\widetilde{Y}_t,X_1-X_0)
\]
is the same as the joint law of
\[
    (X_t,Y_t,X_1-X_0)
\]
at each fixed \(t\).

Let
\[
    D:=X_1-X_0.
\]
The clean bridge satisfies \(dX_t=D\,dt\), so the marginally equivalent process
\(\widetilde{Y}_t=X_t+R_t\) satisfies
\begin{equation}
    d\widetilde{Y}_t
    =
    D\,dt+\sqrt{2\sigma^2t}\,dB_t.
    \label{eq:app_awgn_equiv_sde}
\end{equation}
The drift \(D\) is latent and need not be a deterministic function of
\(\widetilde{Y}_t\).  The one-time marginal forward equation for the density
therefore uses the conditional mean drift
\begin{equation}
    w_t^\star(y)
    :=
    \mathbb{E}[D\mid \widetilde{Y}_t=y,t].
    \label{eq:app_awgn_conditional_drift}
\end{equation}
By the fixed-time joint-law equivalence above,
\begin{equation}
    w_t^\star(y)
    =
    \mathbb{E}[X_1-X_0\mid Y_t=y,t].
    \label{eq:app_noisy_clean_velocity_def}
\end{equation}

The diffusion matrix in \eqref{eq:app_awgn_equiv_sde} is deterministic:
\[
    \bigl(\sqrt{2\sigma^2t}I\bigr)
    \bigl(\sqrt{2\sigma^2t}I\bigr)^\top
    =
    2\sigma^2t\,I.
\]
Hence the one-time marginal forward equation for the density of
\(\widetilde{Y}_t\), and therefore for \(p_t^{\mathrm{noisy}}\), is
\begin{align}
    \partial_t p_t^{\mathrm{noisy}}(y)
    &=
    -
    \nabla_y\cdot
    \left(
        w_t^\star(y)p_t^{\mathrm{noisy}}(y)
    \right)
    +
    \frac12
    \nabla_y\cdot\nabla_y\cdot
    \left(
        2\sigma^2t\,I\,p_t^{\mathrm{noisy}}(y)
    \right)
    \notag \\
    &=
    -
    \nabla_y\cdot
    \left(
        w_t^\star(y)p_t^{\mathrm{noisy}}(y)
    \right)
    +
    \sigma^2t\,\Delta_y p_t^{\mathrm{noisy}}(y).
    \label{eq:app_awgn_transport_diffusion}
\end{align}
This is the desired transport--diffusion form:
\[
    \partial_t p_t^{\mathrm{noisy}}
    =
    -
    \nabla\cdot
    \left(
        w_t^\star p_t^{\mathrm{noisy}}
    \right)
    +
    \sigma^2t\,\Delta p_t^{\mathrm{noisy}}.
\]

For comparison, the clean bridge density satisfies the ordinary clean-flow
continuity equation
\[
    \partial_t p_t^{\mathrm{clean}}
    =
    -
    \nabla\cdot
    \left(
        v_t^{\mathrm{clean}}\, p_t^{\mathrm{clean}}
    \right),
\]
where
\[
    v_t^{\mathrm{clean}}(x)
    =
    \mathbb{E}[X_1-X_0\mid X_t=x,t].
\]
The corresponding observed bridge instead satisfies
\[
    \partial_t p_t^{\mathrm{noisy}}
    =
    -
    \nabla\cdot
    \left(
        v_t^\star p_t^{\mathrm{noisy}}
    \right),
    \qquad
    v_t^\star(y)
    =
    \mathbb{E}[Y-X_0\mid Y_t=y,t].
\]
Indeed, since
\[
    Y-X_0=(X_1-X_0)+\varepsilon_y,
\]
we have
\[
    v_t^\star(y)
    =
    w_t^\star(y)+e_t(y).
\]
For AWGN, Tweedie's identity gives
\[
    e_t(y)p_t^{\mathrm{noisy}}(y)
    =
    -\sigma^2t\,\nabla_y p_t^{\mathrm{noisy}}(y).
\]
Therefore
\[
    -
    \nabla\cdot
    \left(
        v_t^\star p_t^{\mathrm{noisy}}
    \right)
    =
    -
    \nabla\cdot
    \left(
        w_t^\star p_t^{\mathrm{noisy}}
    \right)
    -
    \nabla\cdot
    \left(
        e_t p_t^{\mathrm{noisy}}
    \right)
    =
    -
    \nabla\cdot
    \left(
        w_t^\star p_t^{\mathrm{noisy}}
    \right)
    +
    \sigma^2t\,\Delta p_t^{\mathrm{noisy}}.
\]
Thus the transport--diffusion equation is simply a decomposition of the
ordinary observed-flow continuity equation into the clean posterior drift
\(w_t^\star\) and the divergence contribution of the posterior noise velocity.

The equation describes the marginal density evolution induced by the growing
Gaussian blur \(\mathcal{N}(0,\sigma^2t^2I)\).  

\subsection{Score Identity and AWGN Correction}
\label{app:proof_score_awgn}

We first prove the observed-bridge score identity.  This identity does not use
AWGN; it only uses the Gaussian prior endpoint.  Since
\[
    Y_t=tY+(1-t)X_0,
\]
the Gaussian prior endpoint appears as the additive Gaussian smoothing
component.  Applying the Gaussian Tweedie identity
\eqref{eq:app_tweedie_identity} to the pair
\[
    B=Y_t,
    \qquad
    A=tY,
    \qquad
    \xi=(1-t)X_0,
\]
with covariance \((1-t)^2I\), gives
\begin{equation}
    \mathbb{E}[tY\mid Y_t=y,t]
    =
    y
    +
    (1-t)^2\nabla_y\log p_t^{\mathrm{noisy}}(y).
    \label{eq:app_tY_tweedie}
\end{equation}
Taking conditional expectation in
\[
    Y_t=tY+(1-t)X_0
\]
and substituting \eqref{eq:app_tY_tweedie}, we obtain
\begin{align}
    y
    &=
    \mathbb{E}[tY\mid Y_t=y,t]
    +
    (1-t)\mathbb{E}[X_0\mid Y_t=y,t]
    \notag \\
    &=
    y
    +
    (1-t)^2\nabla_y\log p_t^{\mathrm{noisy}}(y)
    +
    (1-t)\mathbb{E}[X_0\mid Y_t=y,t].
\end{align}
Thus
\begin{equation}
    \mathbb{E}[X_0\mid Y_t=y,t]
    =
    -(1-t)\nabla_y\log p_t^{\mathrm{noisy}}(y).
    \label{eq:app_x0_score}
\end{equation}

Now rewrite the observed bridge as
\[
    Y_t=t(Y-X_0)+X_0.
\]
Taking conditional expectation given \(Y_t=y\),
\begin{equation}
    y
    =
    t\,\mathbb{E}[Y-X_0\mid Y_t=y,t]
    +
    \mathbb{E}[X_0\mid Y_t=y,t].
\end{equation}
By definition of \(v_t^\star\),
\begin{equation}
    \mathbb{E}[X_0\mid Y_t=y,t]
    =
    y-t\,v_t^\star(y).
    \label{eq:app_x0_velocity}
\end{equation}
Combining \eqref{eq:app_x0_score} and \eqref{eq:app_x0_velocity} yields
\begin{equation}
    \nabla_y\log p_t^{\mathrm{noisy}}(y)
    =
    g_t^\star(y)
    =
    \frac{t\,v_t^\star(y)-y}{1-t},
    \qquad t\in(0,1).
    \label{eq:app_score_from_velocity}
\end{equation}
This is \eqref{eq:score_from_velocity}.  The identity is singular at \(t=1\),
which is why the score anchor is evaluated only for \(t<1\).

We next justify the rescaled correction field.  Define
\[
    e_t(y):=\mathbb{E}[\varepsilon_y\mid Y_t=y,t],
    \qquad
    s_t(y):=-\frac{1}{t}e_t(y),
    \qquad t>0.
\]
Since \(Y_t=X_t+t\varepsilon_y\), taking conditional expectation gives
\begin{align}
    \mathbb{E}[X_t\mid Y_t=y,t]
    &=
    y-t\,e_t(y)
    \notag \\
    &=
    y+t^2s_t(y).
    \label{eq:app_clean_state_from_correction}
\end{align}
Similarly,
\[
    Y-X_0=(X_1-X_0)+\varepsilon_y.
\]
Therefore
\begin{align}
    \mathbb{E}[X_1-X_0\mid Y_t=y,t]
    &=
    \mathbb{E}[Y-X_0\mid Y_t=y,t]
    -
    \mathbb{E}[\varepsilon_y\mid Y_t=y,t]
    \notag \\
    &=
    v_t^\star(y)+t\,s_t(y).
    \label{eq:app_clean_velocity_from_correction}
\end{align}
This proves the two posterior-mean identities used in
\eqref{eq:clean_state_velocity_from_correction}.

Now assume AWGN:
\[
    \varepsilon_y\sim\mathcal{N}(0,\sigma^2I),
\]
independent of \((X_0,X_1)\).  Since
\[
    Y_t=X_t+t\varepsilon_y,
\]
the bridge state \(Y_t\) is a Gaussian observation of \(X_t\) with covariance
\(t^2\sigma^2I\).  Applying \eqref{eq:app_tweedie_identity} with
\[
    B=Y_t,
    \qquad
    A=X_t,
    \qquad
    \xi=t\varepsilon_y,
\]
gives
\begin{equation}
    \mathbb{E}[X_t\mid Y_t=y,t]
    =
    y
    +
    t^2\sigma^2\nabla_y\log p_t^{\mathrm{noisy}}(y).
    \label{eq:app_awgn_clean_tweedie}
\end{equation}
Comparing \eqref{eq:app_awgn_clean_tweedie} with
\eqref{eq:app_clean_state_from_correction},
\begin{equation}
    s_t(y)
    =
    \sigma^2\nabla_y\log p_t^{\mathrm{noisy}}(y).
    \label{eq:app_awgn_correction_score}
\end{equation}
Using \eqref{eq:app_score_from_velocity},
\begin{equation}
    s_t(y)
    =
    \sigma^2
    \frac{t\,v_t^\star(y)-y}{1-t}.
    \label{eq:app_awgn_correction_velocity}
\end{equation}
This proves the AWGN correction formula.

Since \(e_t=-t\,s_t\),
\begin{align}
    \mathbb{E}[\varepsilon_y\mid Y_t=y,t]
    &=
    -t\sigma^2\nabla_y\log p_t^{\mathrm{noisy}}(y)
    \notag \\
    &=
    \frac{t\sigma^2}{1-t}
    \left(
        y-t\,v_t^\star(y)
    \right).
    \label{eq:app_awgn_posterior_noise}
\end{align}
The factor \(t\) is necessary: at \(t=0\), the bridge state is \(Y_0=X_0\) and
contains no observation noise.

Finally, using \eqref{eq:app_x0_velocity},
\begin{equation}
    \mathbb{E}[\varepsilon_y\mid Y_t=y,t]
    =
    \frac{t\sigma^2}{1-t}
    \mathbb{E}[X_0\mid Y_t=y,t].
    \label{eq:app_awgn_prior_relation}
\end{equation}

Finally, replacing the population score in
\eqref{eq:app_awgn_correction_score} by the learned score estimate $g_\theta$
from \eqref{eq:score_anchor} gives
$\widehat{s}(y,t)=\sigma^2g_\theta(y,t)$, as stated in
\eqref{eq:awgn_practical_correction}.

\subsection{General Gaussian Correction}
\label{app:proof_hidden_mode_gaussian}

We now derive the correction for general Gaussian corruption. Recall that
$S=\Sigma(X_1,\Omega)$ is the noise covariance for the sample and
$Y_t=X_t+t\varepsilon_y$. Using the bridge notation from
Appendix~\ref{app:proof_preliminaries},
\[
    X_t=(1-t)X_0+tX_1,
    \qquad
    Y_t=X_t+t\varepsilon_y.
\]

For \(t>0\), define the average noise covariance at bridge state $y$ by
\begin{equation}
    \bar{\Sigma}_t(y)
    :=
    \mathbb{E}[S\mid Y_t=y,t].
    \label{eq:app_bar_sigma}
\end{equation}

Conditioned on \((X_0,X_1,\Omega)\), both \(X_t\) and \(S\) are fixed, and
\begin{equation}
    Y_t\mid X_0,X_1,\Omega
    \sim
    \mathcal{N}(X_t,t^2S).
    \label{eq:app_hidden_conditional_path}
\end{equation}
Let \(\ell_t(y\mid X_0,X_1,\Omega)\) denote this conditional Gaussian density.
Differentiating it with respect to \(y\) gives
\begin{equation}
    t^2S
    \nabla_y
    \ell_t(y\mid X_0,X_1,\Omega)
    =
    -(y-X_t)
    \ell_t(y\mid X_0,X_1,\Omega).
    \label{eq:app_hidden_likelihood_gradient}
\end{equation}

The posterior numerator of the scaled observation noise can therefore be
written as
\begin{align}
    p_t^{\mathrm{noisy}}(y)
    \mathbb{E}[t\varepsilon_y\mid Y_t=y,t]
    &=
    \mathbb{E}
    \left[
        (y-X_t)
        \ell_t(y\mid X_0,X_1,\Omega)
    \right]
    \notag \\
    &=
    -\operatorname{div}_y
    \mathbb{E}
    \left[
        t^2S
        \ell_t(y\mid X_0,X_1,\Omega)
    \right].
    \label{eq:app_hidden_noise_numerator}
\end{align}

By Bayes' rule,
\begin{equation}
    \mathbb{E}
    \left[
        S\ell_t(y\mid X_0,X_1,\Omega)
    \right]
    =
    p_t^{\mathrm{noisy}}(y)\bar{\Sigma}_t(y).
    \label{eq:app_hidden_bayes_covariance}
\end{equation}

Substituting this identity into \eqref{eq:app_hidden_noise_numerator} gives
\begin{equation}
    p_t^{\mathrm{noisy}}(y)
    \mathbb{E}[t\varepsilon_y\mid Y_t=y,t]
    =
    -t^2
    \operatorname{div}_y
    \left(
        p_t^{\mathrm{noisy}}(y)\bar{\Sigma}_t(y)
    \right).
    \label{eq:app_hidden_scaled_noise_divergence}
\end{equation}

Using the product rule from \eqref{eq:app_matrix_product_rule} and dividing by
\(p_t^{\mathrm{noisy}}(y)\), we obtain
\begin{equation}
    \mathbb{E}[t\varepsilon_y\mid Y_t=y,t]
    =
    -t^2
    \left[
        \bar{\Sigma}_t(y)
        \nabla_y\log p_t^{\mathrm{noisy}}(y)
        +
        \operatorname{div}\bar{\Sigma}_t(y)
    \right].
    \label{eq:app_hidden_t_epsilon}
\end{equation}

Equivalently, the posterior mean observation noise is
\begin{equation}
    e_t(y)
    :=
    \mathbb{E}[\varepsilon_y\mid Y_t=y,t]
    =
    -t
    \left[
        \bar{\Sigma}_t(y)
        \nabla_y\log p_t^{\mathrm{noisy}}(y)
        +
        \operatorname{div}\bar{\Sigma}_t(y)
    \right].
    \label{eq:app_hidden_posterior_noise}
\end{equation}

Since \(s_t(y)=-e_t(y)/t\), the correction field is
\begin{equation}
    s_t(y)
    =
    \bar{\Sigma}_t(y)
    \nabla_y\log p_t^{\mathrm{noisy}}(y)
    +
    \operatorname{div}\bar{\Sigma}_t(y).
    \label{eq:app_hidden_correction}
\end{equation}
This is the general Gaussian correction in
\eqref{eq:general_correction_field}. Using the score
identity from \eqref{eq:app_score_from_velocity}, the same correction can be
written as
\begin{equation}
    s_t(y)
    =
    \bar{\Sigma}_t(y)g_t^\star(y)
    +
    \operatorname{div}\bar{\Sigma}_t(y).
    \label{eq:app_hidden_correction_velocity}
\end{equation}
for \(t\in(0,1)\).

Finally, substituting \(e_t(y)=-t s_t(y)\) into
\(Y_t=X_t+t\varepsilon_y\) gives the clean-state posterior mean
\begin{equation}
    \mathbb{E}[X_t\mid Y_t=y,t]
    =
    y+t^2s_t(y).
    \label{eq:app_hidden_clean_state}
\end{equation}

When \(S=\sigma^2I\), its posterior expectation is constant,
\[
    \bar{\Sigma}_t(y)=\sigma^2I,
\]
and the divergence term vanishes.  The expression in
\eqref{eq:app_hidden_correction} then reduces to the AWGN correction from
Appendix~\ref{app:proof_score_awgn}.

The general Gaussian correction replaces the fixed AWGN covariance
$\sigma^2I$ with the average noise covariance at the current bridge state and
adds the divergence term that appears when this covariance varies with the
state.

\subsection{Auxiliary Correction Objective}
\label{app:proof_auxiliary_correction}

The general Gaussian correction depends on the average noise covariance at the
current bridge state and is not directly available. The auxiliary objective
learns it by adding a small Gaussian perturbation whose covariance is based on
the noise covariance $S$ for the same training sample. The correction is
\[
    s_t(y)
    =
    \bar{\Sigma}_t(y)\nabla_y\log p_t^{\mathrm{noisy}}(y)
    +
    \operatorname{div}\bar{\Sigma}_t(y)
\]
and requires averages over clean images and sample noise settings.

Let
\[
    S=\Sigma(X_1,\Omega)
\]
be the noise covariance for the training sample.
Draw \(\eta\sim\mathcal{N}(0,I)\) independently and define
\[
    \Delta_{t,\delta}
    =
    \sqrt{\delta}\,t\,S^{1/2}\eta,
    \qquad
    \widetilde{Y}_{t,\kappa,\delta}
    =
    Y_t+\kappa\Delta_{t,\delta},
\]
with \(\delta>0\), \(\kappa\neq0\), and \(t>0\).  Conditional on \(S\), the
auxiliary perturbation \(\kappa\Delta_{t,\delta}\) has covariance
\[
    \kappa^2\delta t^2S.
\]

Applying the same calculation as in
Appendix~\ref{app:proof_hidden_mode_gaussian} to
\(\widetilde{Y}_{t,\kappa,\delta}=Y_t+\kappa\Delta_{t,\delta}\), whose
perturbation covariance for fixed $S$ is \(\kappa^2\delta t^2S\), gives the
corresponding correction for the additionally smoothed bridge.
Let \(p_{t,\kappa,\delta}\) be the density of
\(\widetilde{Y}_{t,\kappa,\delta}\), and define
\begin{equation}
    \bar S_{t,\kappa,\delta}(\widetilde y)
    :=
    \mathbb{E}
    \left[
        S\mid
        \widetilde{Y}_{t,\kappa,\delta}=\widetilde y,t
    \right].
    \label{eq:app_aux_bar_s}
\end{equation}
This calculation gives
\begin{align}
    \mathbb{E}
    \left[
        \kappa\Delta_{t,\delta}
        \mid
        \widetilde{Y}_{t,\kappa,\delta}=\widetilde y,t
    \right]
    &=
    -\kappa^2\delta t^2
    \left[
        \bar S_{t,\kappa,\delta}(\widetilde y)
        \nabla_{\widetilde y}
        \log p_{t,\kappa,\delta}(\widetilde y)
        +
        \operatorname{div}
        \bar S_{t,\kappa,\delta}(\widetilde y)
    \right].
    \label{eq:app_aux_kappa_delta_mean}
\end{align}
Define the smoothed auxiliary correction field
\begin{equation}
    s_{t,\kappa,\delta}(\widetilde y)
    :=
    \bar S_{t,\kappa,\delta}(\widetilde y)
    \nabla_{\widetilde y}\log p_{t,\kappa,\delta}(\widetilde y)
    +
    \operatorname{div}
    \bar S_{t,\kappa,\delta}(\widetilde y).
    \label{eq:app_aux_smoothed_correction}
\end{equation}
Dividing \eqref{eq:app_aux_kappa_delta_mean} by \(\kappa\), we obtain
\begin{equation}
    \mathbb{E}
    \left[
        \Delta_{t,\delta}
        \mid
        \widetilde{Y}_{t,\kappa,\delta}=\widetilde y,t
    \right]
    =
    -\kappa\delta t^2
    s_{t,\kappa,\delta}(\widetilde y).
    \label{eq:app_aux_delta_mean}
\end{equation}

Now consider the auxiliary loss
\[
    \mathcal{L}_{\mathrm{aux}}(\phi)
    =
    \mathbb{E}
    \left[
        \left\|
        -\kappa\delta t^2
        s_\phi(\widetilde{Y}_{t,\kappa,\delta},t)
        -
        \Delta_{t,\delta}
        \right\|^2
    \right].
\]
If \(\kappa\), \(\delta\), and \(t\) are fixed, then for each fixed
\(\widetilde y\), the conditional least-squares minimizer must satisfy
\[
    -\kappa\delta t^2s_\phi(\widetilde y,t)
    =
    \mathbb{E}
    \left[
        \Delta_{t,\delta}
        \mid
        \widetilde{Y}_{t,\kappa,\delta}=\widetilde y,t
    \right].
\]
Using \eqref{eq:app_aux_delta_mean}, the finite-\(\delta\) population minimizer
is
\begin{equation}
    s_\phi(\widetilde y,t)
    =
    s_{t,\kappa,\delta}(\widetilde y).
    \label{eq:app_aux_finite_delta_target}
\end{equation}
Thus, at finite \(\delta\), the objective identifies the correction field of the
additionally smoothed bridge, not exactly the correction field of the original
observed bridge.

To recover the desired field, we take the small-perturbation limit.  We assume
that as \(\delta\rightarrow0\),
\[
    p_{t,\kappa,\delta}\to p_t^{\mathrm{noisy}},
    \qquad
    \bar S_{t,\kappa,\delta}\to\bar{\Sigma}_t,
\]
and that their first spatial derivatives converge locally wherever
\(p_t^{\mathrm{noisy}}>0\).  Under this \(C^1\)-type convergence,
\begin{equation}
    s_{t,\kappa,\delta}(\widetilde y)
    \to
    s_t(\widetilde y)
    \qquad
    \text{as }
    \delta\rightarrow0.
    \label{eq:app_aux_limit}
\end{equation}
Equivalently,
\begin{equation}
    -
    \frac{1}{\kappa\delta t^2}
    \mathbb{E}
    \left[
        \Delta_{t,\delta}
        \mid
        \widetilde{Y}_{t,\kappa,\delta}=\widetilde y,t
    \right]
    \to
    s_t(\widetilde y).
    \label{eq:app_aux_rescaled_limit}
\end{equation}

If \(\kappa\) or \(\delta\) is randomized during training and not provided as an
input to \(s_\phi\), then the finite-perturbation population minimizer is the
corresponding conditional average over perturbation levels.  Let
\[
    a_{\kappa,\delta,t}:=\kappa\delta t^2.
\]
The pointwise minimizer is
\begin{equation}
    s_\phi(\widetilde y,t)
    =
    \frac{
        \mathbb{E}
        \left[
            a_{\kappa,\delta,t}^2
            s_{t,\kappa,\delta}(\widetilde y)
            \mid
            \widetilde{Y}_{t,\kappa,\delta}=\widetilde y,t
        \right]
    }{
        \mathbb{E}
        \left[
            a_{\kappa,\delta,t}^2
            \mid
            \widetilde{Y}_{t,\kappa,\delta}=\widetilde y,t
        \right]
    }.
    \label{eq:app_aux_randomized_minimizer}
\end{equation}
In the small-\(\delta\) regime, this distinction vanishes under the same
smooth-convergence assumptions.

Finally, the noise covariance \(S=\Sigma(X_1,\Omega)\) for the sample must be
reused in the auxiliary corruption. If an independent covariance \(S'\) were
sampled instead,
then
\[
    \widetilde{Y}_{t,\kappa,\delta}\to Y_t
    \qquad
    \text{as }
    \delta\rightarrow0,
\]
but \(S'\) would remain independent of \(Y_t\).  Consequently,
\[
    \mathbb{E}
    \left[
        S'\mid
        \widetilde{Y}_{t,\kappa,\delta}=\widetilde y,t
    \right]
    \to
    \mathbb{E}[S'],
\]
rather than
\[
    \mathbb{E}[S\mid Y_t=\widetilde y,t]
    =
    \bar{\Sigma}_t(\widetilde y).
\]
The learned correction would then collapse to a population-average covariance
correction and would lose the state-dependent covariance information.

\section{Ablations}\label{subsec:ablations}

\subsection{Perturbation Range}
\label{app:ablation_delta_kappa}

\begin{table}[htbp]
\centering
\caption{Ablation of the $\delta$ and $\kappa$ hyperparameters. Bold denotes the best result and underline denotes results within 1 FID.}
\label{tab:results}
\scriptsize
\begin{tabular}{ccccc}
\toprule
$\delta_{\min}$ & $\delta_{\max}$ & $\kappa_{\min}$ & $\kappa_{\max}$ & FID \\
\midrule
0.001 & 0.1 & 0   & 1 & \underline{19.459} \\
0.005 & 0.1 & 0   & 1 & \underline{18.974} \\
0.01  & 0.1 & 0   & 1 & \textbf{18.869} \\
\cmidrule(lr){1-5}
0.001 & 0.1 & 0.5 & 1 & \underline{18.971} \\
0.005 & 0.1 & 0.5 & 1 & \underline{19.113} \\
0.01  & 0.1 & 0.5 & 1 & \underline{19.443} \\
\midrule
0.001 & 0.1 & 0   & 2 & 29.064 \\
0.005 & 0.1 & 0   & 2 & 28.029 \\
0.01  & 0.1 & 0   & 2 & 28.445 \\
\cmidrule(lr){1-5}
0.001 & 0.1 & 0.5 & 2 & 28.067 \\
0.005 & 0.1 & 0.5 & 2 & 28.297 \\
0.01  & 0.1 & 0.5 & 2 & 27.094 \\
\midrule
0.001 & 0.3 & 0   & 1 & 23.591 \\
0.005 & 0.3 & 0   & 1 & 23.703 \\
0.01  & 0.3 & 0   & 1 & 23.361 \\
\cmidrule(lr){1-5}
0.001 & 0.3 & 0.5 & 1 & 24.710 \\
0.005 & 0.3 & 0.5 & 1 & 24.422 \\
0.01  & 0.3 & 0.5 & 1 & 24.782 \\
\midrule
0.001 & 0.3 & 0   & 2 & 62.026 \\
0.005 & 0.3 & 0   & 2 & 63.673 \\
0.01  & 0.3 & 0   & 2 & 65.275 \\
\cmidrule(lr){1-5}
0.001 & 0.3 & 0.5 & 2 & 64.001 \\
0.005 & 0.3 & 0.5 & 2 & 65.646 \\
0.01  & 0.3 & 0.5 & 2 & 67.217 \\
\bottomrule
\end{tabular}
\end{table}

Table~\ref{tab:results} reports an ablation of the perturbation ranges on
CelebA-HQ at $32\times32$ resolution under variable AWGN with
$\sigma\in[0.04,0.4]$.  The upper bounds $\delta_{\max}$ and
$\kappa_{\max}$ have the largest effect on performance and should both be kept
relatively small.  Increasing $\delta_{\max}$ from $0.1$ to $0.3$ consistently
degrades FID, while increasing $\kappa_{\max}$ from $1$ to $2$ causes an even
larger deterioration, especially when combined with the larger
$\delta_{\max}$.  In contrast, varying $\delta_{\min}$ or $\kappa_{\min}$
within the tested ranges has only a minor effect.
The best result is obtained with $\delta_{\min}=0.01$,
$\delta_{\max}=0.1$, $\kappa_{\min}=0$, and $\kappa_{\max}=1$.

\subsection{Sampling Cutoff}
\label{app:ablation_sampling_cutoff}

\begin{table}[htbp]
\centering
\caption{FID ($\downarrow$) ablation of the sampling cutoff
$t_{\mathrm{cut}}$ on CelebA-HQ and CIFAR-10 under AWGN. Bold denotes the best
result for each dataset and noise level.}
\label{tab:t_cutoff_ablation}
\scriptsize
\setlength{\tabcolsep}{5pt}
\begin{tabular}{ccccccc}
\toprule
& \multicolumn{3}{c}{CelebA-HQ} & \multicolumn{3}{c}{CIFAR-10} \\
\cmidrule(lr){2-4}\cmidrule(lr){5-7}
$t_{\mathrm{cut}}$
& $\sigma=0.05$ & $\sigma=0.10$ & $\sigma=0.20$
& $\sigma=0.05$ & $\sigma=0.10$ & $\sigma=0.20$ \\
\midrule
0.950 & 8.4163 & 10.6508 & 12.9096 & 7.9085 & 10.3194 & 13.1515 \\
0.960 & 7.9051 & 10.3132 & 11.3524 & 7.5676 & 10.1812 & 12.0904 \\
0.970 & 7.5972 & 10.1223 & 10.3395 & 7.2422 & 9.9371 & 9.7197 \\
0.975 & 7.4778 & 10.0685 & \textbf{10.0621} & 7.1066 & 9.6798 & \textbf{9.0423} \\
0.980 & 7.2019 & 9.8535 & 10.1460 & 7.0054 & 9.5695 & 9.7703 \\
0.985 & 6.7707 & 9.0468 & 13.3111 & 6.8833 & 8.7999 & 14.6737 \\
0.990 & 6.8979 & \textbf{6.8526} & 26.3845 & 6.7769 & \textbf{6.2115} & 34.3116 \\
0.995 & \textbf{6.4994} & 13.9124 & 55.5894 & \textbf{6.4276} & 15.0627 & 85.3487 \\
\bottomrule
\end{tabular}
\end{table}

\begin{table}[htbp]
\centering
\caption{Extension of Table~\ref{tab:ambient-awgn} with NR-CFM evaluated using
the FID-optimal cutoff for each noise level.  The selected cutoffs are
$t_{\mathrm{cut}}=0.995$, $0.990$, and $0.975$ for
$\sigma=0.05$, $0.10$, and $0.20$, respectively.  Bold denotes the best result
and underline denotes results within 1 FID.  Values are mean $\pm$ standard
deviation over three seeds.}
\label{tab:ambient_awgn_optimal_cutoff}
\small
\setlength{\tabcolsep}{3pt}
\begin{tabular}{lccc}
\toprule
\multicolumn{4}{c}{CIFAR-10} \\
\cmidrule(lr){1-4}
Method & $\sigma=0.05$ & $\sigma=0.10$ & $\sigma=0.20$ \\
\midrule
CFM & $44.55\pm0.12$ & $82.98\pm0.08$ & $135.98\pm0.18$ \\
Ambient Diffusion & $\mathbf{2.82\pm0.02}$ & $\mathbf{3.63\pm0.03}$
& $11.93\pm0.09$ \\
NR-CFM ($t_{\mathrm{cut}}=0.98$) & $7.00\pm0.04$ & $9.51\pm0.04$
& $\underline{9.71\pm0.12}$ \\
NR-CFM (optimal $t_{\mathrm{cut}}$) & $6.37\pm0.04$ & $6.16\pm0.03$
& $\mathbf{8.94\pm0.10}$ \\
\midrule
\multicolumn{4}{c}{CelebA-HQ} \\
\cmidrule(lr){1-4}
Method & $\sigma=0.05$ & $\sigma=0.10$ & $\sigma=0.20$ \\
\midrule
CFM & $40.09\pm0.15$ & $62.81\pm0.02$ & $108.32\pm0.08$ \\
Ambient Diffusion & $\mathbf{5.50\pm0.03}$ & $9.38\pm0.02$
& $12.97\pm0.11$ \\
NR-CFM ($t_{\mathrm{cut}}=0.98$) & $7.15\pm0.03$ & $9.88\pm0.05$
& $\underline{10.09\pm0.09}$ \\
NR-CFM (optimal $t_{\mathrm{cut}}$) & $\underline{6.45\pm0.03}$
& $\mathbf{6.89\pm0.02}$ & $\mathbf{10.02\pm0.08}$ \\
\bottomrule
\end{tabular}
\end{table}

Table~\ref{tab:t_cutoff_ablation} shows that sampling performance is sensitive
to the choice of $t_{\mathrm{cut}}$, particularly at higher noise levels.  The
preferred cutoff decreases as the noise standard deviation increases: the best
results use $t_{\mathrm{cut}}=0.995$ for $\sigma=0.05$, $0.990$ for
$\sigma=0.10$, and $0.975$ for $\sigma=0.20$.  At larger $\sigma$, extending
the correction too close to the endpoint leads to a sharp degradation in FID,
possibly because the score function has a larger magnitude and therefore
induces stronger, less stable corrections near $t=1$.  Notably, the optimal
cutoff at each noise level is identical for CelebA-HQ and CIFAR-10, suggesting
that its dependence on the corruption strength is consistent across the two
datasets.  Based on this asymmetric sensitivity, we use the conservative
choice $t_{\mathrm{cut}}=0.95$ for the diverse NR-GAN settings and the
extreme-noise CryoBench experiments: stopping somewhat early generally incurs
only a modest loss, whereas stopping too late can cause a much larger
degradation.  For the more controlled Ambient Diffusion comparison, we use
$t_{\mathrm{cut}}=0.98$, which provides a favorable balance across AWGN levels
and yields competitive performance without tuning the cutoff separately for
each noise strength.

Table~\ref{tab:ambient_awgn_optimal_cutoff} quantifies the potential benefit of
per-noise cutoff selection.  Relative to the fixed $t_{\mathrm{cut}}=0.98$
used in the main comparison, the optimal cutoff improves FID substantially in
several settings, most notably at $\sigma=0.10$.  However, these cutoffs are
selected by evaluating FID against clean reference data and should therefore be
viewed as an oracle ablation rather than a deployable model-selection
procedure.  In the intended noisy-data-only setting, clean samples are not
available for choosing $t_{\mathrm{cut}}$.  Reducing the sensitivity of NR-CFM
to this hyperparameter, or developing a criterion that selects it using only
corrupted observations, is an important direction for future work.

\section{Additional Generated Image Examples}
\label{app:generated_examples}

\begin{figure}[p]
    \centering
    \includegraphics[width=0.95\linewidth]{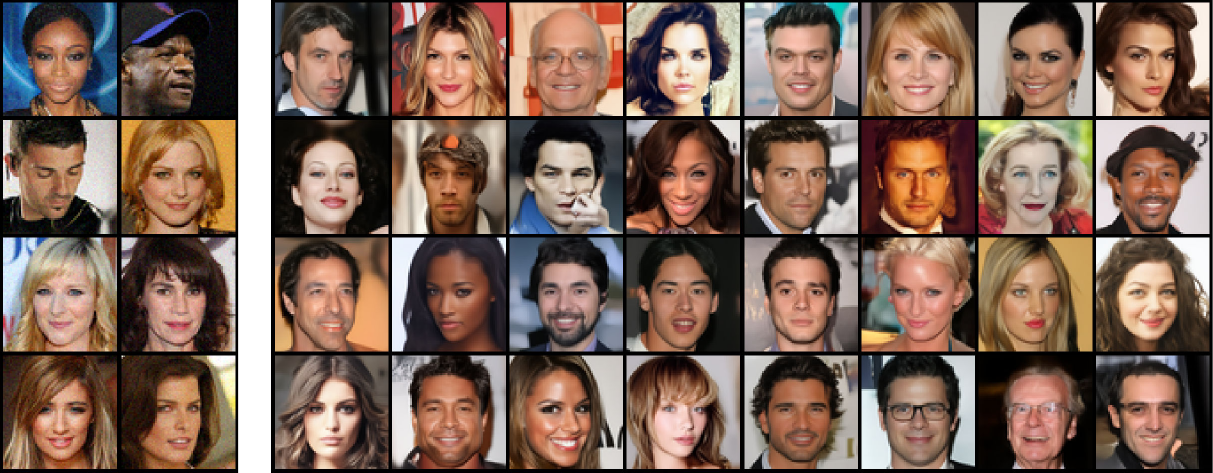}
    \vspace{8pt}

    \includegraphics[width=0.95\linewidth]{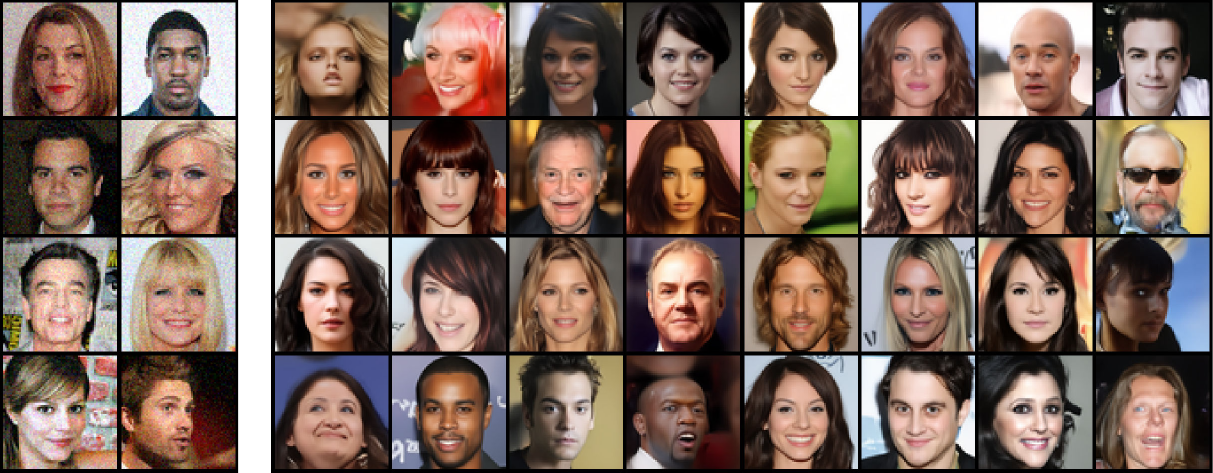}
    \vspace{8pt}

    \includegraphics[width=0.95\linewidth]{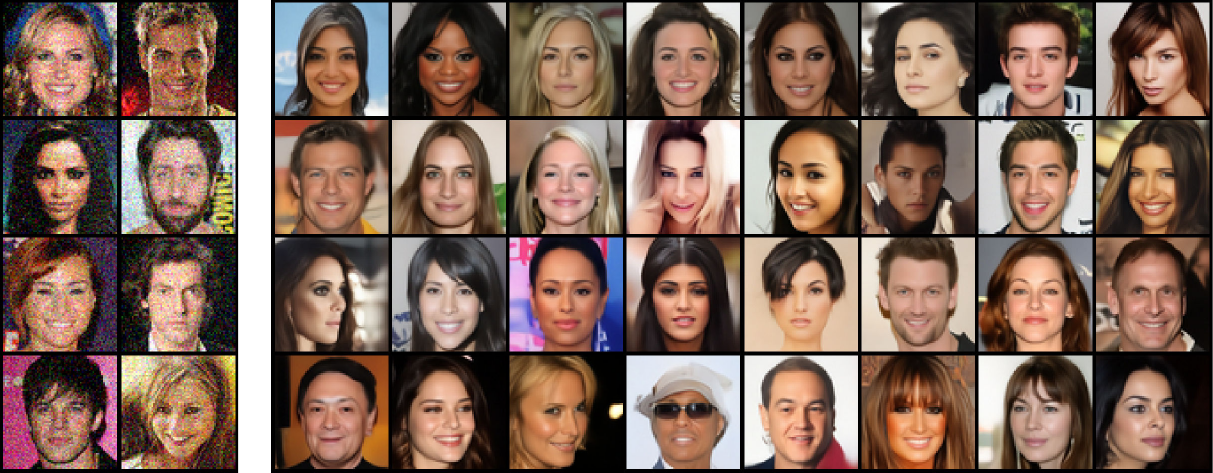}
    \caption{CelebA-HQ examples under AWGN with, from top to bottom,
    $\sigma=0.05$, $\sigma=0.1$, and $\sigma=0.2$.
    }
    \label{fig:appendix_celeba_awgn}
\end{figure}

\begin{figure}[p]
    \centering
    \includegraphics[width=0.95\linewidth]{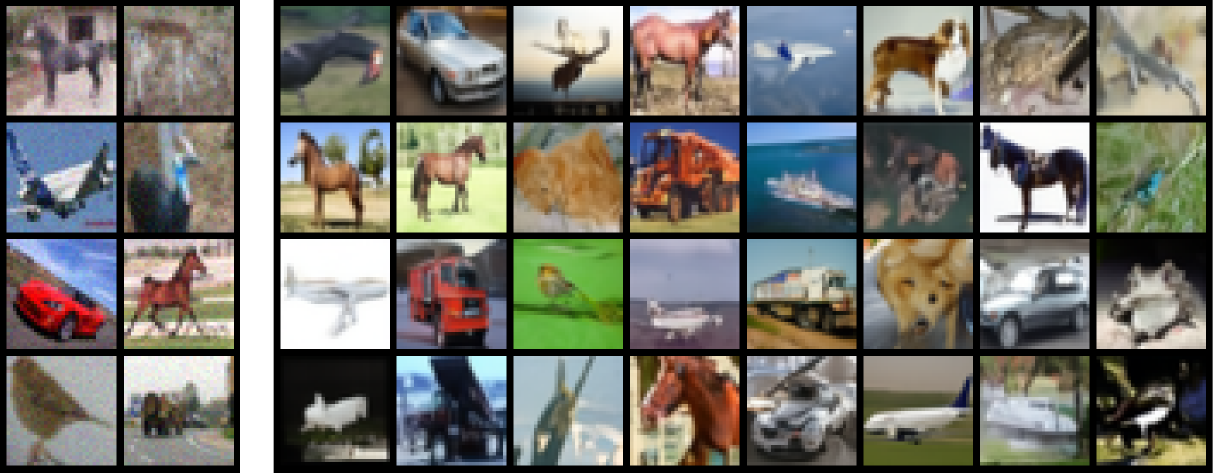}
    \vspace{8pt}

    \includegraphics[width=0.95\linewidth]{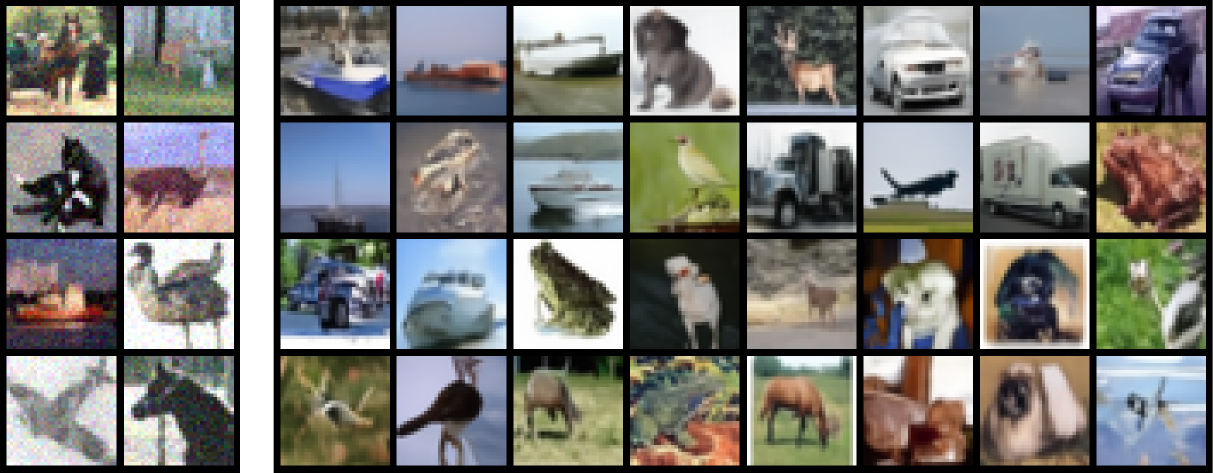}
    \vspace{8pt}

    \includegraphics[width=0.95\linewidth]{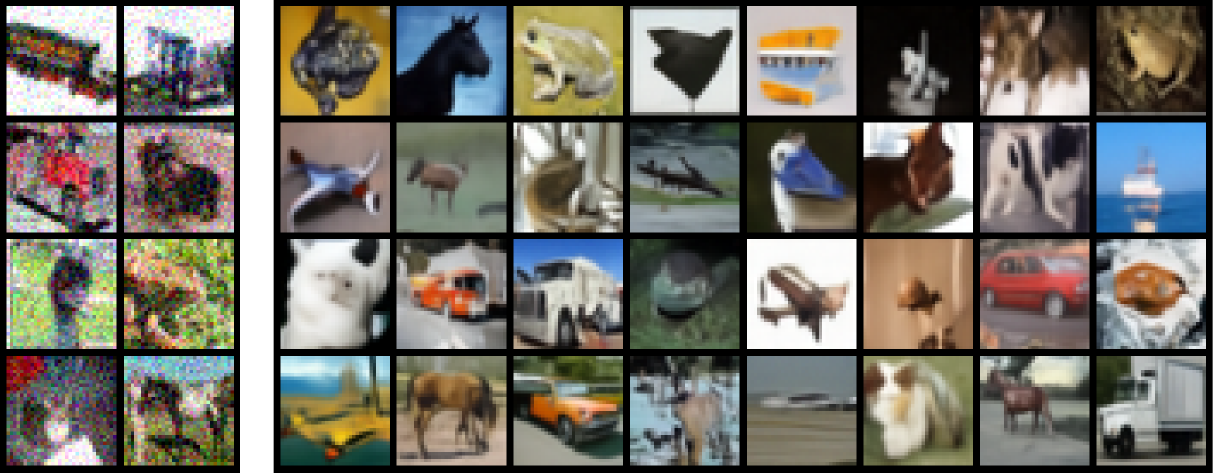}
    \caption{CIFAR-10 examples under AWGN with, from top to bottom,
    $\sigma=0.05$, $\sigma=0.1$, and $\sigma=0.2$.
    The final setting also corresponds to NR-GAN experiment (A).
    }
    \label{fig:appendix_cifar_awgn}
\end{figure}

\begin{figure}[p]
    \centering
    \includegraphics[width=0.95\linewidth]{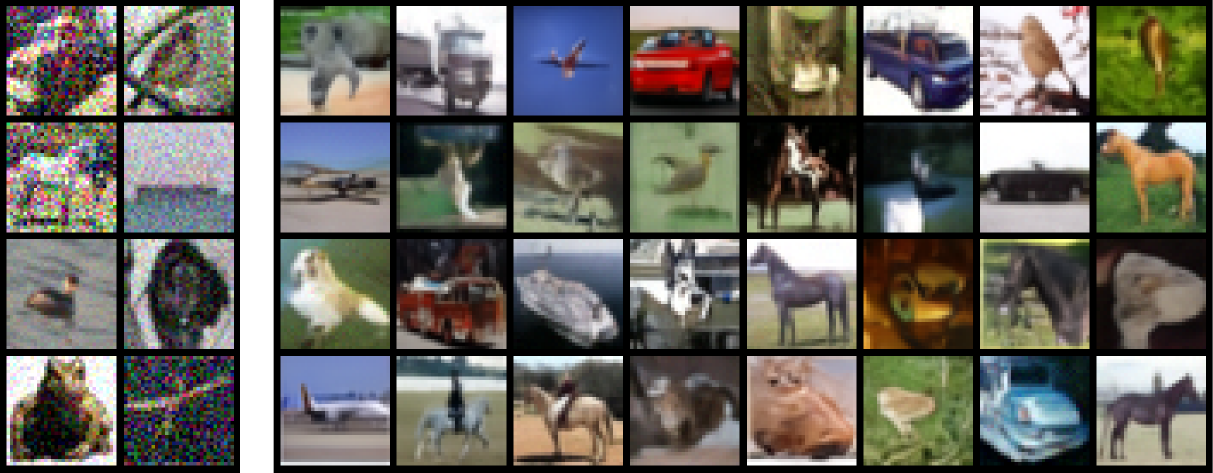}
    \vspace{8pt}

    \includegraphics[width=0.95\linewidth]{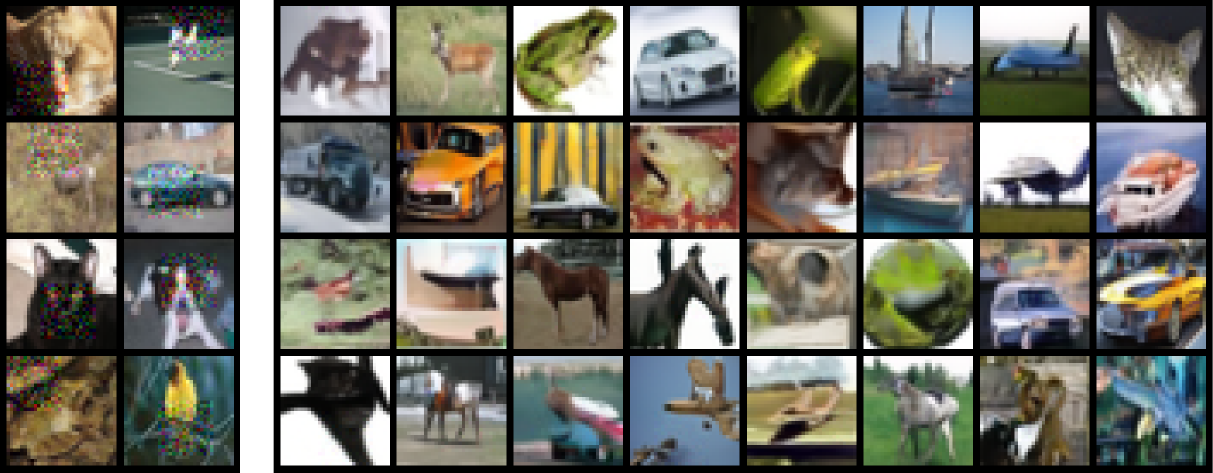}
    \vspace{8pt}

    \includegraphics[width=0.95\linewidth]{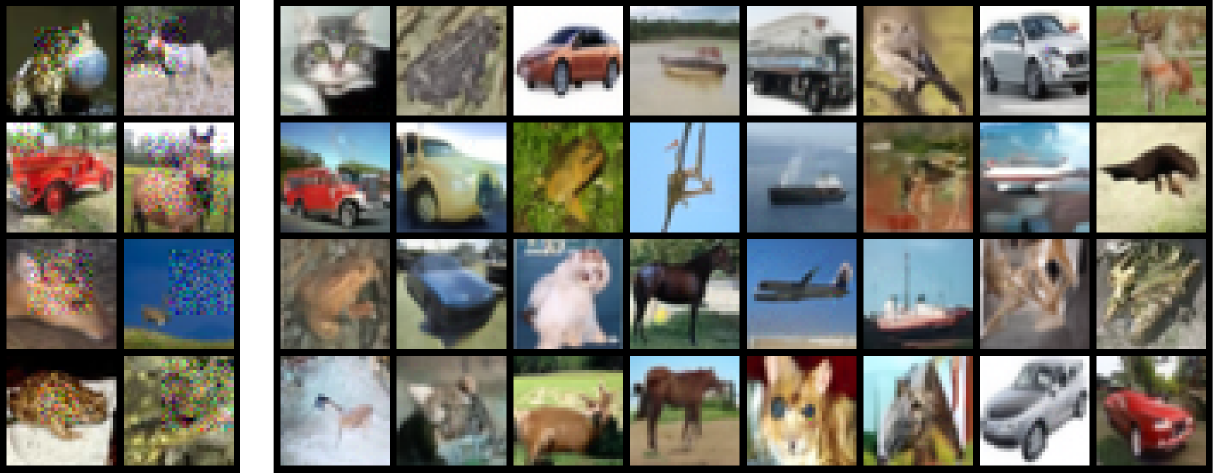}
    \caption{CIFAR-10 examples for NR-GAN settings B--D, from top to bottom:
    additive Gaussian noise with variable magnitude (B), local Gaussian noise
    on a fixed patch (C), and local Gaussian noise on a variable patch (D).}
    \label{fig:appendix_cifar_bd}
\end{figure}

\begin{figure}[p]
    \centering
    \includegraphics[width=0.95\linewidth]{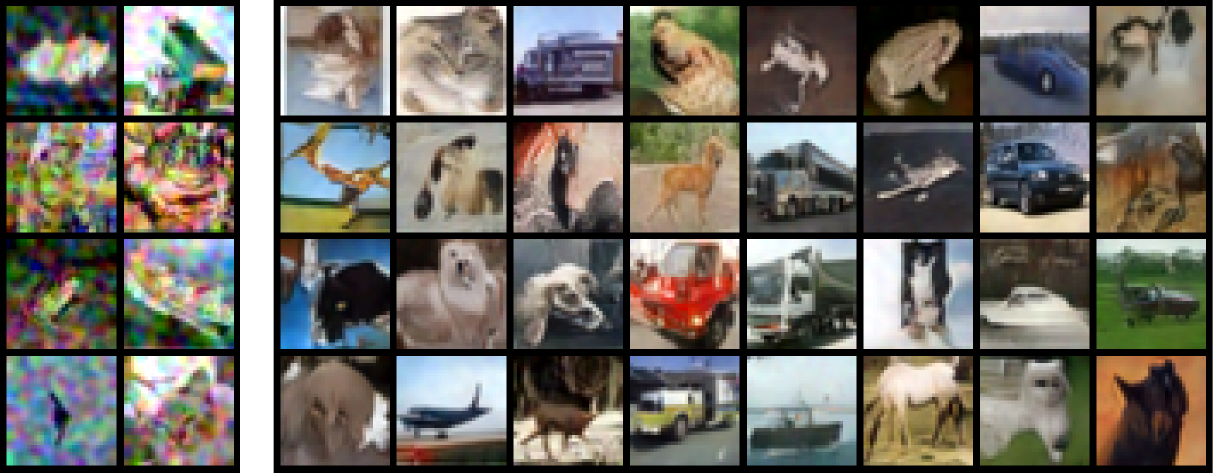}
    \vspace{8pt}

    \includegraphics[width=0.95\linewidth]{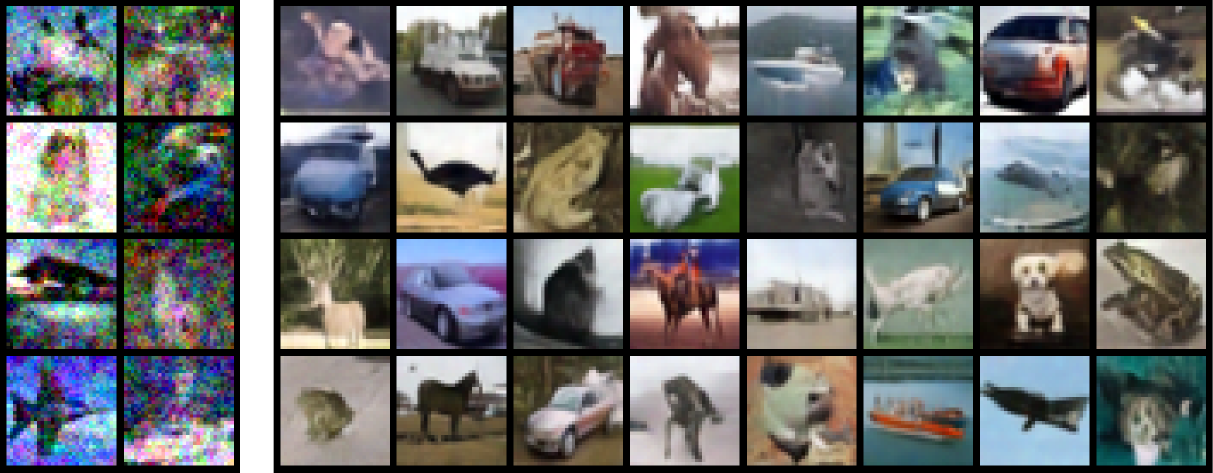}
    \vspace{8pt}

    \includegraphics[width=0.95\linewidth]{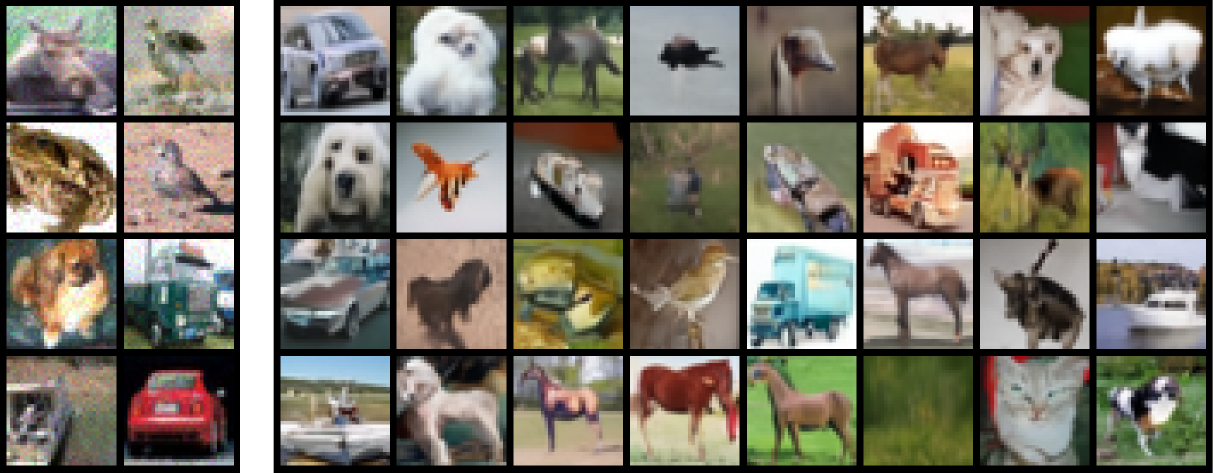}
    \caption{CIFAR-10 examples for NR-GAN settings G--I, from top to bottom:
    Brown Gaussian noise (G), additive white plus Brown Gaussian noise (H),
    and multiplicative Gaussian noise with fixed magnitude (I).}
    \label{fig:appendix_cifar_gi}
\end{figure}

\begin{figure}[p]
    \centering
    \includegraphics[width=0.95\linewidth]{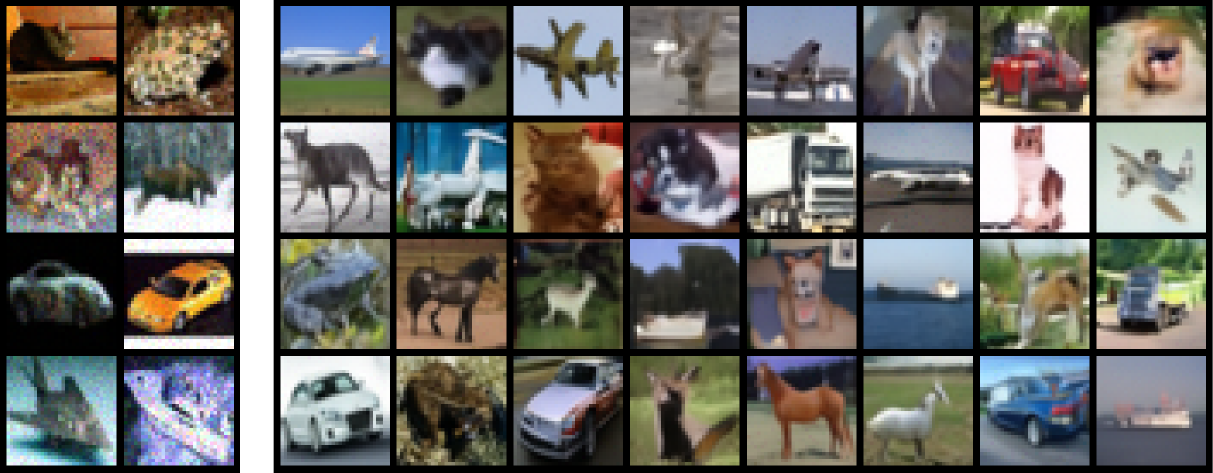}
    \vspace{8pt}

    \includegraphics[width=0.95\linewidth]{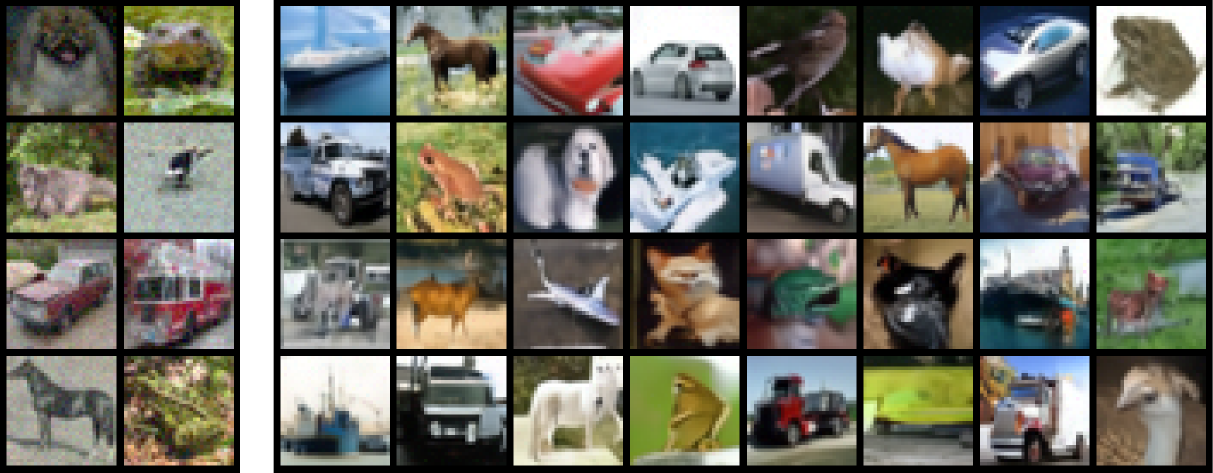}
    \vspace{8pt}

    \includegraphics[width=0.95\linewidth]{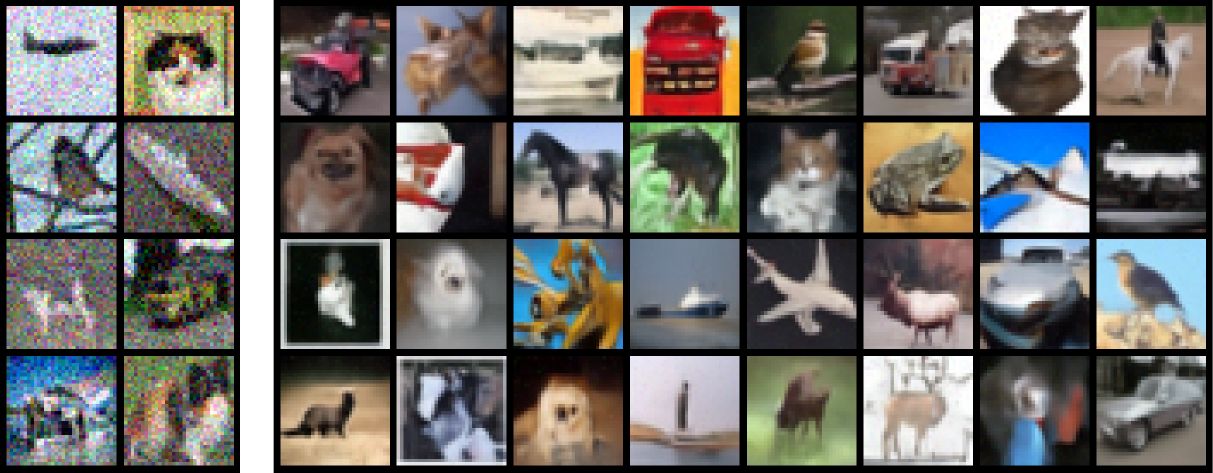}
    \caption{CIFAR-10 examples for NR-GAN settings J--L, from top to bottom:
    multiplicative Gaussian noise with variable magnitude (J) and additive
    plus multiplicative Gaussian noise at the two tested magnitudes (K and L).}
    \label{fig:appendix_cifar_jl}
\end{figure}

\begin{figure}[p]
    \centering
    \includegraphics[width=0.95\linewidth]{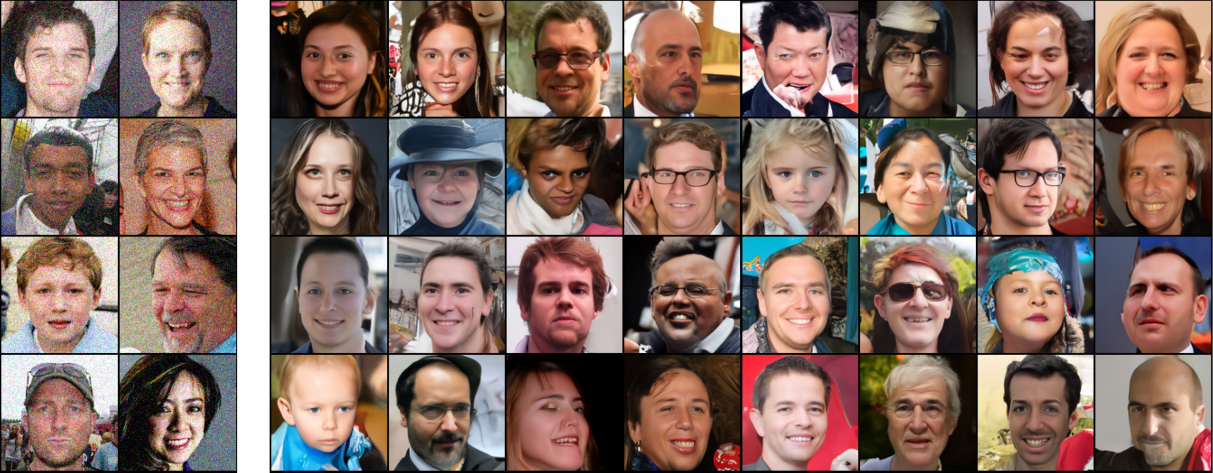}
    \vspace{8pt}

    \includegraphics[width=0.95\linewidth]{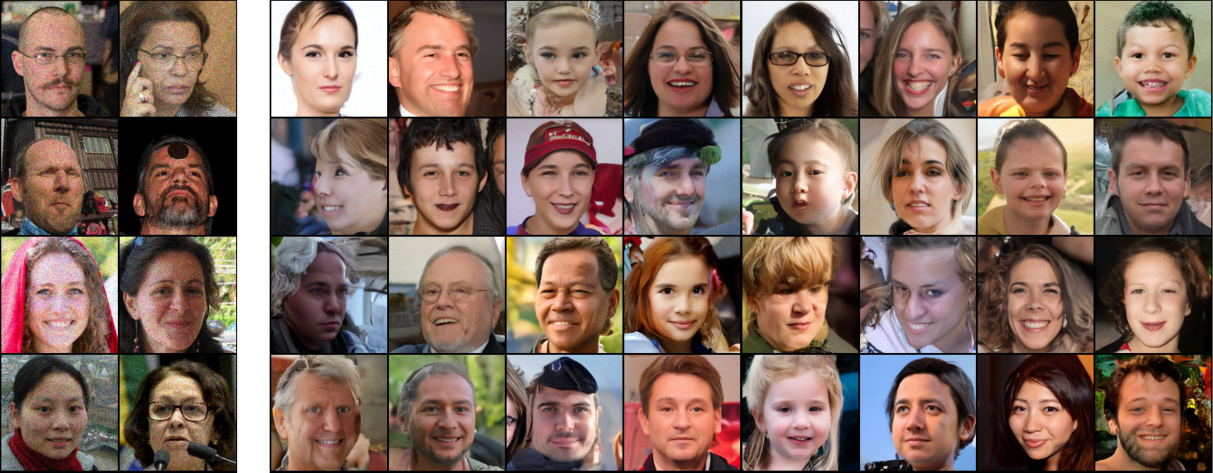}
    \caption{FFHQ examples for NR-GAN settings A and I, from top to bottom:
    additive Gaussian noise with fixed magnitude (A) and multiplicative
    Gaussian noise with fixed magnitude (I).}
    \label{fig:appendix_ffhq_ai}
\end{figure}

\begin{figure}[p]
    \centering
    \includegraphics[width=0.95\linewidth]{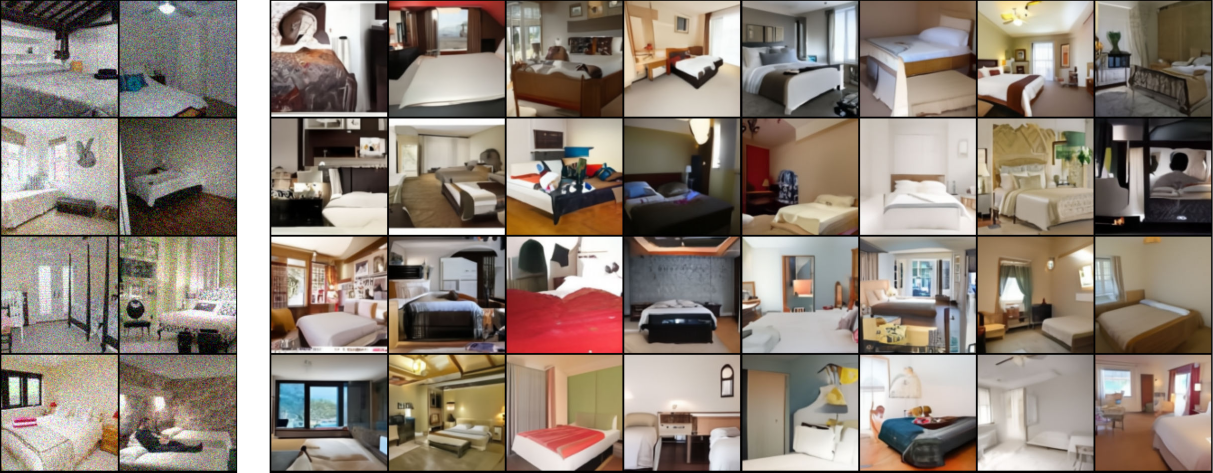}
    \vspace{8pt}

    \includegraphics[width=0.95\linewidth]{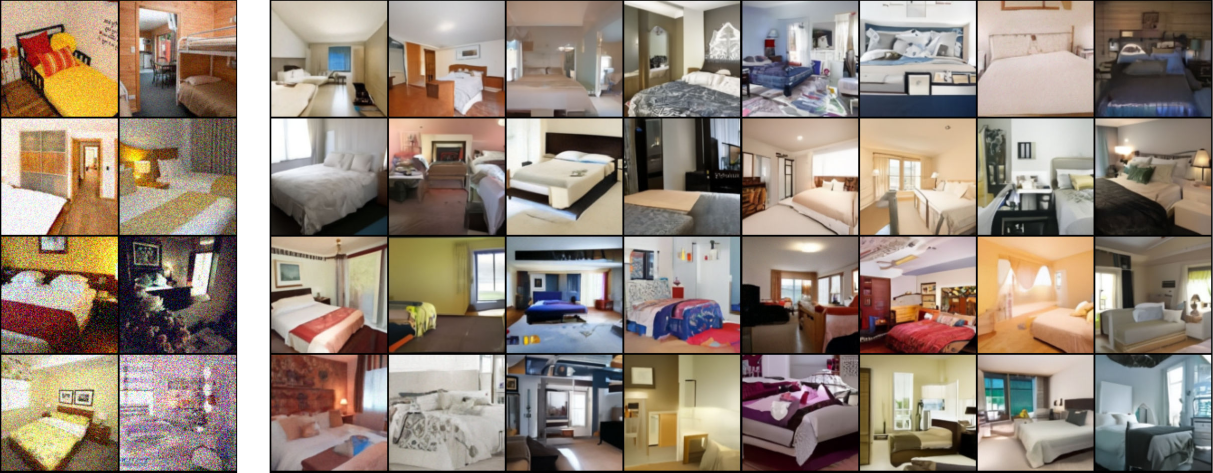}
    \vspace{8pt}

    \includegraphics[width=0.95\linewidth]{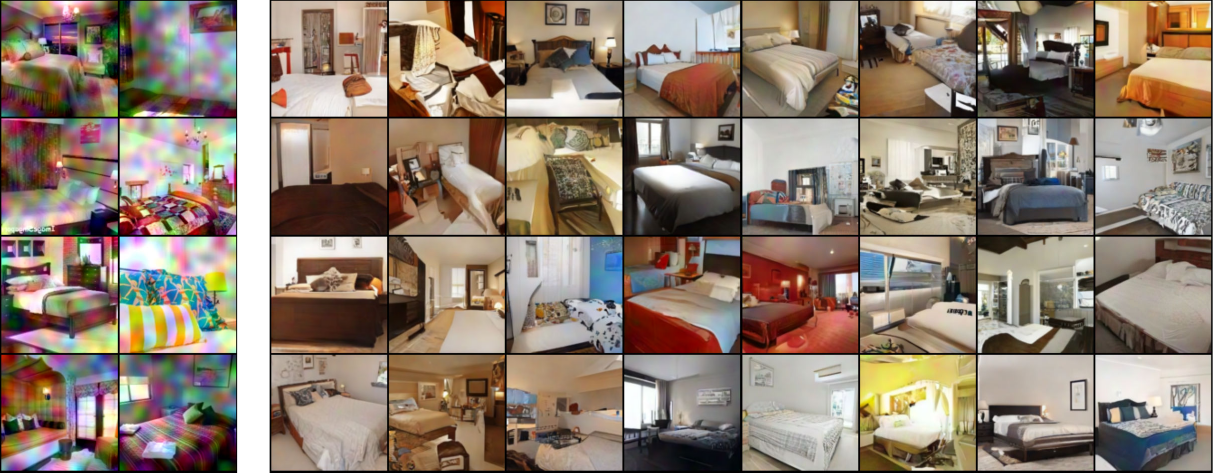}
    \caption{LSUN Bedroom examples for NR-GAN settings A, B, and G, from top
    to bottom: additive Gaussian noise with fixed magnitude (A), additive
    Gaussian noise with variable magnitude (B), and Brown Gaussian noise (G).}
    \label{fig:appendix_lsun_abg}
\end{figure}

\begin{figure}[p]
    \centering
    \includegraphics[width=0.95\linewidth]{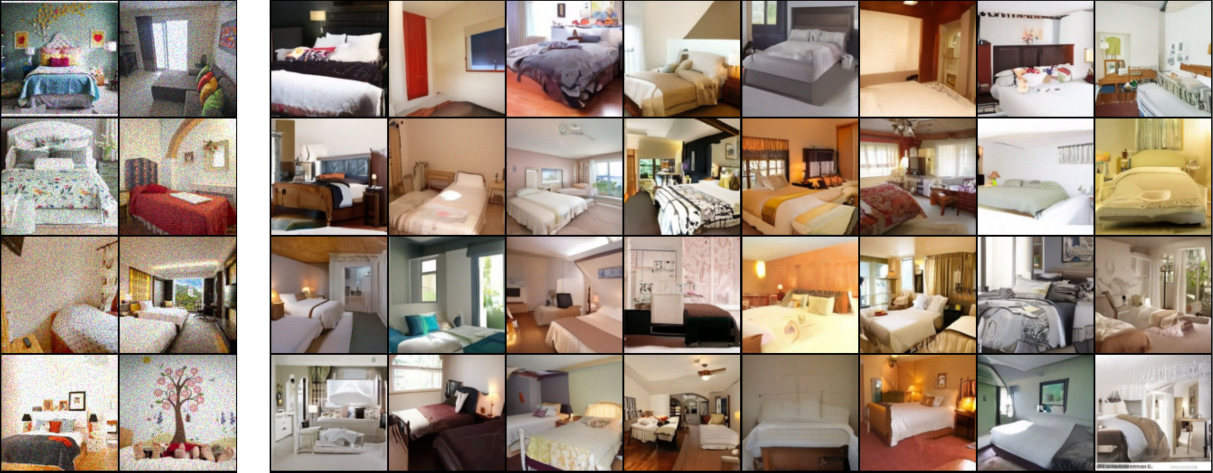}
    \vspace{8pt}

    \includegraphics[width=0.95\linewidth]{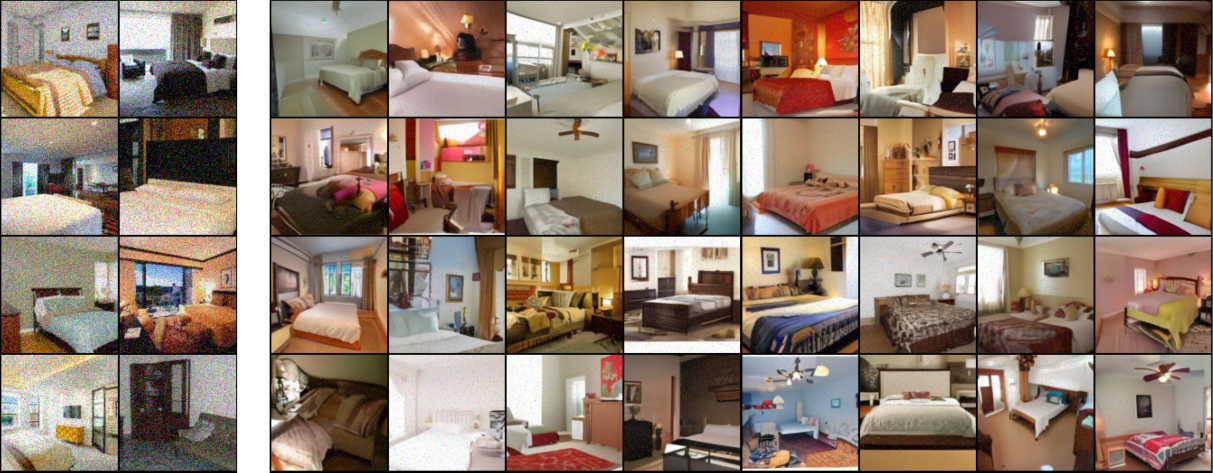}
    \caption{LSUN Bedroom examples for NR-GAN settings I and L, from top to
    bottom: multiplicative Gaussian noise with fixed magnitude (I) and
    additive plus multiplicative Gaussian noise (L).}
    \label{fig:appendix_lsun_il}
\end{figure}

Figures~\ref{fig:appendix_celeba_awgn}--\ref{fig:appendix_lsun_il} show
additional generated samples across the datasets and corruption settings used
in our experiments.  In every panel, the left half contains example corrupted
source images and the right half contains images generated by NR-CFM.  These
examples complement the quantitative results by illustrating generation
quality across a broad range of Gaussian corruption regimes.

\newpage

\end{document}